\documentclass[12pt]{article}
\usepackage{amsmath}
\usepackage{times}
\usepackage{graphicx}
\usepackage{color}
\usepackage{multirow}
\usepackage{rotating}
\usepackage{bbm}
\usepackage{latexsym}

\usepackage{marvosym}
\usepackage{lineno,hyperref}
\usepackage{array}
\usepackage{booktabs}
\usepackage{caption}
\usepackage{mathtools} 
\modulolinenumbers[5]
\newcommand{\ve}[1]{\mathbf{#1}}

\newcommand{\captionfonts}{\normalsize}

\makeatletter  
\long\def\@makecaption#1#2{%
  \vskip\abovecaptionskip
  \sbox\@tempboxa{{\captionfonts #1: #2}}%
  \ifdim \wd\@tempboxa >\hsize
    {\captionfonts #1: #2\par}
  \else
    \hbox to\hsize{\hfil\box\@tempboxa\hfil}%
  \fi
  \vskip\belowcaptionskip}
\makeatother   

\begin{document}
\hspace{13.9cm}

\ \\
{\LARGE Continual Learning with Bayesian Model based on a Fixed Pre-trained Feature Extractor}

\ \\
{\bf \large Yang Yang$^{\displaystyle 1, \displaystyle 2}$, \large Zhiying Cui$^{\displaystyle 1, \displaystyle 2}$, \large Junjie Xu$^{\displaystyle 1, \displaystyle 2}$, \large Changhong Zhong$^{\displaystyle 1, \displaystyle 2}$, \large Wei-Shi Zheng$^{\displaystyle 1, \displaystyle 2}$, \large Ruixuan Wang$^{\displaystyle 1, \displaystyle 2*}$}\\ 
{$^{\displaystyle 1}$School of Computer Science and Engineering, Sun Yat-sen University, China.}\\
{$^{\displaystyle 2}$Key Laboratory of Machine Intelligence and Advanced Computing, MOE,  China.}\\
%


\thispagestyle{empty}
\markboth{}{NC instructions}
%
%
\begin{center} {\bf Abstract} \end{center}
Deep learning has shown its human-level performance in various applications. However, current deep learning models are characterised by catastrophic forgetting of old knowledge when learning new classes. This poses a challenge particularly in intelligent diagnosis systems where initially only training data of a limited number of diseases are available. In this case, updating the intelligent system with data of new diseases would inevitably downgrade its performance on previously learned diseases. 
Inspired by the process of learning new knowledge in human brains, we propose a Bayesian generative model for continual learning built on a fixed pre-trained feature extractor. In this model, knowledge of each old class can be compactly represented by a collection of statistical distributions, e.g. with Gaussian mixture models, 
and naturally kept from forgetting in continual learning over time. Unlike existing class-incremental learning methods, the proposed approach is not sensitive to the continual learning process and can be additionally well applied to the data-incremental learning scenario.
Experiments on multiple medical and natural image classification tasks showed that the proposed approach outperforms state-of-the-art approaches which even keep some images of old classes during continual learning of new classes.

\section{Introduction}
Deep learning models, particularly convolutional neural networks (CNNs), have shown human-level performance in various applications, such as in healthcare~\cite{ardila2019end,de2018clinically,esteva2017dermatologist,mckinney2020international}, surveillance~\cite{ciaparrone2020deep,sam2020locate,xiong2019open,hu2018sinet}, and machine translation~\cite{radford2019language,vaswani2017attention}. However, particularly in the healthcare domain, most intelligent diagnosis systems are limited to diagnosis of only one or a few diseases and cannot be easily extended once deployed, and therefore cannot 
diagnose all diseases of certain tissue or organ (e.g., skin or lung) as medical specialists do. Since collecting data of all (e.g., skin or lung) diseases is challenging due to various reasons (e.g., privacy and limited data sharing), it is impractical to train an intelligent system diagnosing all diseases at once. One possible solution is to make the intelligent system have the continual or lifelong learning ability, such that it can continually learn to diagnose more and more diseases without resourcing (or resourcing few) original data of previously learned diseases~\cite{baweja2018towards}. Such continual learning of new classes may also appear in other applications such as in automated retail stores~\cite{diethe2019continual}.
However, current intelligent models are characterised by catastrophic forgetting of old knowledge when learning new classes~\cite{french1999catastrophic,goodfellow2013empirical,kemker2018measuring}.

Researchers have recently proposed multiple types of continual learning approaches to reduce  catastrophic forgetting of old knowledge particularly in deep learning models~\cite{kirkpatrick2017overcoming,yoon2018lifelong,li2017learning,rebuffi2017icarl,shin2017continual}. The overall objective is to help the updated classifier accurately recognize both new and old classes, when only data of new classes and few (or even no) data of old classes are available during classifier updating. 
However, almost all existing approaches modify the feature extraction part of the classifiers either in parameter values or in structures during continual learning of new classes. In contrast, humans seem to learn new knowledge by adding memory of the learned new information without modifying the (e.g., visual) perceptual pathway. Therefore, one possible cause to catastrophic forgetting in existing models is the change in the feature extraction part (corresponding to the perceptual pathway in human brains) when learning new knowledge. With this consideration, we propose a generative model for continual learning built on a fixed pre-trained feature extractor, which is different from all existing (discriminative) models. The generative model can naturally keep knowledge of each old class from forgetting, without storing original images of old classes or regenerating synthetic images during continual learning. Experiments on two skin disease classification tasks and two natural image classification tasks showed the proposed approach outperforms state-of-the-art approaches which even keep some images of old classes during continual learning. The proposed approach provides a new direction for investigation of continual learning, i.e., exploring effective ways to represent and store knowledge of each class based on a fixed but powerful feature extractor.

\section{Related Work}

There are typically two types of continual learning problems, task-incremental and class-incremental. Task-incremental learning presumes that one model is incrementally updated to solve more and more tasks, often with multiple tasks sharing a common feature extractor but having task-specific classification heads. The task identification is presumed to be available during inference, i.e., users know which classification head should be applied when predicting the class label of a new test data. This setting is impractical for intelligent diagnosis systems where old and new diseases need to be diagnosed together. In contrast, class incremental learning presumes that one model is incrementally updated to predict more and more classes sharing a single classification head. This approach is more relevant to the continual learning of new diseases. Thus, our study focuses on the class-incremental learning problem. Existing approaches to the two types of continual learning can be roughly divided into four groups, regularization-based, expansion-based, distillation-based and regeneration-based.

Regularization-based approaches often estimate model components (e.g. kernels in CNNs) crucial for old knowledge, and try to change them as little as possible with the help of regularization loss terms when learning new knowledge~\cite{kirkpatrick2017overcoming,abati2020conditional,ahn2019uncertainty,fernando2017pathnet,jung2020continual,kim2018keep,mallya2018packnet,zenke2017continual}. The importance of each model parameter can be measured by the sensitivity of the loss function to changes in model parameter as in the elastic weight consolidation (EWC) method~\cite{kirkpatrick2017overcoming}, or by the sensitivity of the model output to small changes in model parameter as in the memory aware synapses (MAS) method~\cite{aljundi2018memory}. The importance of each kernel in a CNN model can be measured based on the magnitude of the kernel (e.g., L2 norm of the kernel matrix) as in the PackNet~\cite{mallya2018packnet}. 
Regularization-based approaches could help models keep old knowledge at first few rounds of continual learning where no much new knowledge need to be learned. However, it would become more and more difficult to continually learn new knowledge particularly at later rounds of continual learning because more and more kernels in CNNs become crucial and therefore should be kept unchanged for increasing old knowledge.

To make models more flexibly learn new knowledge, expansion-based approaches are developed to modify model structures by adding new kernels, layers, or even sub-networks when learning new knowledge~\cite{yoon2018lifelong,aljundi2017expert,hung2019compacting,karani2018lifelong,li2019learn,rajasegaran2019random,verma2021efficient,yan2021dynamically,douillard2021dytox}. For example, Aljundi et al.~\cite{aljundi2017expert} proposed employing an additional network for a new task and training an expert model to make decisions about which network to use during inference. It turns a class-incremental learning problem to a task-incremental problem at the cost of additional parameters. As another example, Yoon et al. proposed a dynamically expandable network (DEN)~\cite{yoon2018lifelong} by selectively retraining the network and expanding kernels at each layer if necessary. Most expansion-based and regularization-based approaches are initially proposed for task-incremental learning, although some of them (e.g., EWC) can be extended for class-incremental learning.

In comparison, distillation-based approaches can be directly applied to continual learning of new classes by distilling knowledge from the old classifier (for old classes) to the new classifier (for both new and old classes) during learning new knowledge~\cite{li2017learning,rebuffi2017icarl,iscen2020memory,castro2018end,hou2018lifelong,meng2020adinet,hu2021distilling,pourkeshavarzi2021looking,smith2021always}, where the old knowledge is often implicitly represented by soft outputs of old classifier with stored small amount of old images and/or new classes of images as the inputs. A distillation loss is added to the original cross-entropy loss during training the new classifier, where the distillation loss helps the new classifier have similar relevant output compared to the output of the old classifier for any input image. The well-known methods include the learning without forgetting (LwF)~\cite{li2017learning}, the incremental classifier and representation learning (iCaRL)~\cite{rebuffi2017icarl}, and the end-to-end incremental learning (End2End)~\cite{castro2018end}. More recently, the distillation has been extended to intermediate CNN layers, either by keeping feature map activation unchanged as in the learning without memorizing (LwM)~\cite{dhar2019learning}, or by keeping the spatial pooling unchanged respectively along the horizontal and vertical directions as in PODNet~\cite{douillard2020podnet}, or by keeping the normalized global pooling unchanged at last convolutional layers as in learning a unified classifier incrementally via rebalancing (UCIR)~\cite{hou2019learning}. These distillation-based methods achieve state-of-the-art performance for the class-incremental learning problem. However, such methods would become insufficient with continual learning of more classes, either because the number of stored old data becomes too small to be representative for each old class, or because the outputs of the old classifier with new classes of data as inputs cannot represent knowledge of old classes due to underlying differences between new classes and each old class.

In addition, regeneration-based approaches have also been proposed particularly when none of the old-data is available during learning new classes. The basic idea is to train an auto-encoder~\cite{hayes2019remind,rao2019continual,riemer2019scalable} or generative adversarial network (GAN)~\cite{shin2017continual,ostapenko2019learning,rios2018closed,xiang2019incremental} to produce enough number of realistic data for each old class when learning new classes. The potential issue is that fine-grained lesion features may not be well learned by the generative model, which would result in unsatisfying synthetic data when updating the intelligent diagnosis system. Different from all the existing approaches, we propose a simple but effective generative model which is based on a fixed pre-trained feature extractor and does not store any old data.

\section{A Generative Model for Continual Learning}\label{sec4}
The proposed method is inspired by two interesting findings in neuroscience. One finding is that most infants cannot form episodic memory before 3 years old~\cite{bauer2015complementary,ribordy2013development,scarf2013have}, and the other finding is that humans continually form memory from infants to elderly people~\cite{nadel2012memory}. One hypothetical explanation is that the visual pathway in younger infant's brain might be rapidly changing with daily visual stimuli from surroundings and then become firm with little change since 3 years old or so. Humans can continually learn new visual knowledge through their whole lives probably because they form new memories about the new knowledge, but without changing the visual pathway which works as a visual feature extractor. This could help explain why current deep learning models are characterised by catastrophic forgetting of old knowledge, i.e., model parameters or model structures from the feature extractor part are always changed to some extent in almost all continual learning approaches. With this consideration, we propose a human-like continual learning framework, i.e., first pre-training a feature extractor, then fixing the feature extractor and forming new memory for new knowledge. In the following, we will introduce one general way to pre-train the feature extractor, one statistical method to represent the memory, and one Bayesian model to predict class of any new (test) data after continual learning each time.

\subsection{Fixed pre-trained feature extractor}
\begin{figure*}[!tbh]
    \centering
    \includegraphics[width=1\linewidth]{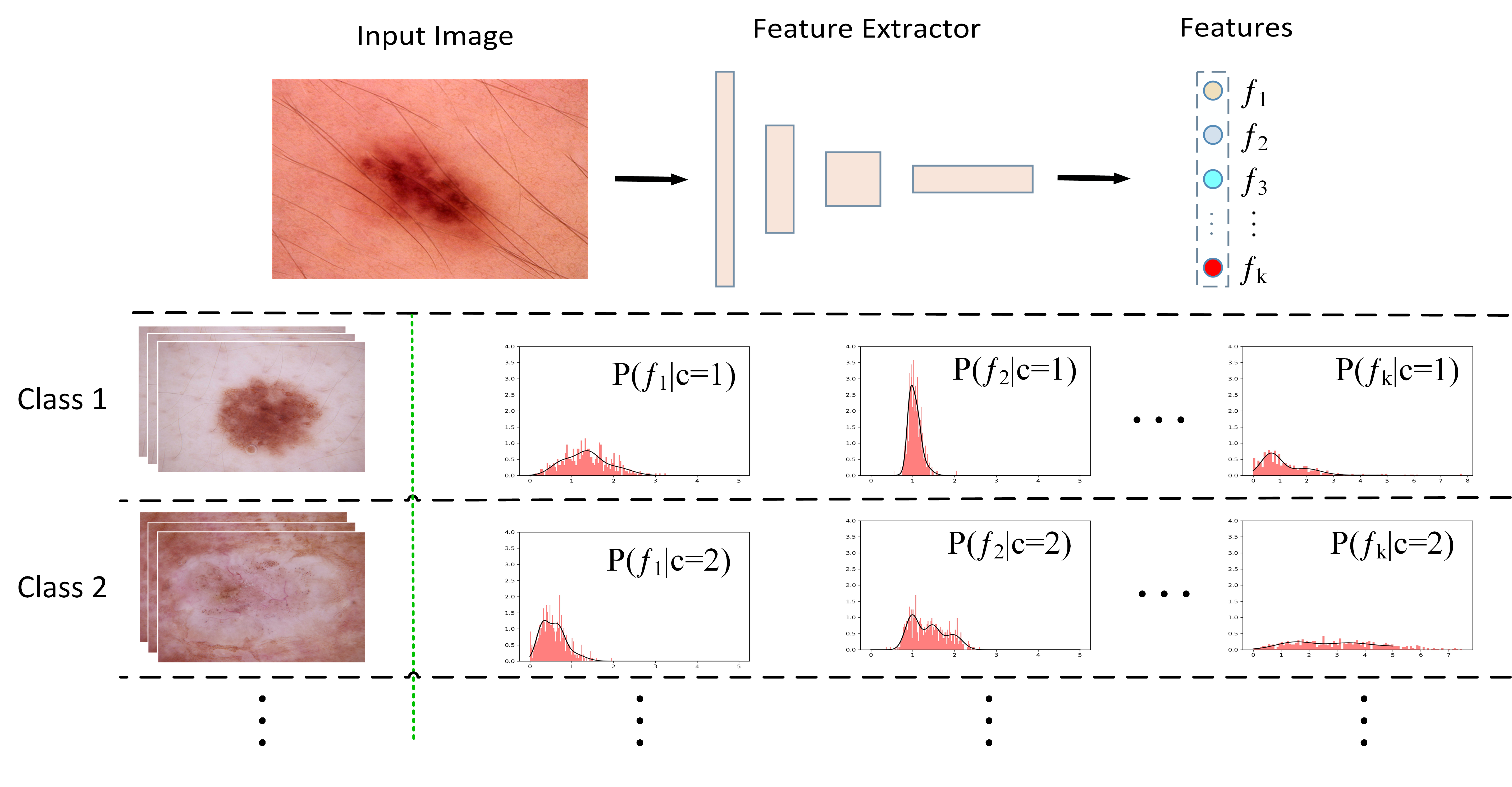}    
    \caption{Fixed pre-trained Feature extractor (top) and memory formation (middle to bottom). Feature extractor is pre-trained and fixed during continual learning. Memory of each class is represented by a set of statistical distributions over features.}
    \label{fig:framework}
\end{figure*}

An ideal feature extractor should output two different feature vectors if two input data were visually different, meanwhile visually more similar inputs should result in more similar feature vectors from the feature extractor. The visual feature extractor (i.e., visual pathway) in younger infants are probably taught in certain self-supervised way, although the mechanism of self-supervision in infant brain has not been explicitly understood~\cite{ribordy2013development}. While it is worth exploring various self-supervised learning approaches (e.g., auto-encoder) to train a feature extractor, here we leave the self-supervision exploration for future work, and adopt a simpler but widely used approach, i.e., pre-training a CNN classifier with relatively large number images whose classes or domains are relevant but different from those in the task of interest, and then using the pre-trained CNN feature extractor (often consisting of all the convolutoinal layers;  Figure~\ref{fig:framework}, top row) for the continual learning classification task of interest. 
It is expected that the pre-trained feature extractor would probably be powerful enough to discriminate different input images in the task of interest.
Experiments in this study show that even such a simple approach to a fixed pre-trained feature extractor can already significantly help reduce catastrophic forgetting of old knowledge with the proposed generative approach.
It is worth noting that, during continual learning of new classes in the classification task of interest, the pre-trained feature extractor is fixed and not updated. The knowledge of each learned new class is represented and stored as described in the following subsection.

\subsection{Memory formation}
Different from state-of-the-art continual learning approaches which often store a small number of original images for each old class, the proposed approach stores not original images but the statistical information of each class based on the feature extractor outputs of all training images belonging to the class. Here, each element of the output feature vector is assumed to represent certain type of visual feature. Then based on the class of training images, the distribution of each feature is estimated and collected together to form the memory of the knowledge of the specific class (Figure~\ref{fig:framework}, second and third rows, each row for one class). Formally, denote by $D_c=\{\ve{x}_i, i=1,\ldots,N_c\}$ the set of training images for class $c$, $\ve{z}_i = [z_{i1}, z_{i2},\ldots,z_{ik},\ldots,z_{iK}]^{\mathsf{T}}$ the $L_2$-normalized output feature vector from the feature extractor for the input image $\ve{x}_i$, and $\ve{f} = [f_{1}, f_{2},\ldots,f_{k},\ldots,f_{K}]^{\mathsf{T}}$ the vector of random variables representing the output of the feature extractor, then the statistical distribution of the $k$-th feature $f_k$ for class $c$ can be represented by a probability density distribution $p(f_k \vert c, D_c)$,
\begin{equation} \label{eq:likelihood}
p(f_k \vert c, D_c)= g(\{z_{ik}, i=1,\ldots,N_c\}) \,, \,\,\quad \forall \, k \in \{1,\ldots, K\}
\end{equation}
where $g(\cdot)$ could be any appropriate distribution estimator. Here a Gaussian mixture model (GMM) with a small number of $S$ components is adopted to represent $g(\cdot)$ for its simplicity. Since each Gaussian component can be compactly represented by its mean and standard deviation, totally only $2 \cdot S \cdot K$ numbers are stored in the memory to represent the knowledge of each class. $D_c$ would be omitted from $p(f_k \vert c, D_c)$ in the following for simplicity. 

\subsection{Bayesian model for prediction}
Based on the statistical distributions of visual features for each class, we propose a generative classification model with the Bayesian rule for prediction. Given a test image $\ve{x}_j$, denote by $\ve{z}_j = [z_{j1}, z_{j2},\ldots,z_{jk},\ldots,z_{jK}]^{\mathsf{T}}$ the corresponding output from the feature extractor, and $p(c \vert \ve{z}_j)$ the probability of the test image belonging to class $c$.  Then based on the Bayes rule, we can get
\begin{equation} \label{eq:bayes}
p(c \vert \ve{z}_j)= \frac{p(\ve{z}_j \vert c)\cdot p(c)}{\sum^{M}_{m=1} p(\ve{z}_j \vert m)\cdot p(m)} \,,
\end{equation}
where $M$ is the number of classes learned so far. Considering that potential correlations between certain feature components are probably caused by co-occurred visual parts of a specific class of objects, it can be assumed that different feature components $f_k$'s are conditionally independent given specific class $c$. Then, the logarithm of Equation~\ref{eq:bayes} gives
\begin{equation} \label{eq:bayes_log}
\log p(c \vert \ve{z}_j)= \sum_k \log p(f_k=z_{jk} \vert c) + \log p(c) - \alpha \,,
\end{equation}
where $\alpha = \log \sum_m p(\ve{z}_j \vert m) p(m)$ can be considered a constant for different classes. In Equation~(\ref{eq:bayes_log}), the likelihood function value $p(f_k=z_{jk} \vert c)$ for each feature element $k$ can be directly obtained based on the previously stored knowledge $p(f_k \vert c)$ (Equation~\ref{eq:likelihood}) in the memory. The prior $p(c)$ for class $c$ can be simply estimated based on the ratio of the number of training images for this class over the total number of training images of all learned classes so far, i.e., $p(c) = N_c/\sum_m N_m$. Note that in this case, the number of training images for each class needs to be stored in the memory such that $p(c)$ can be easily updated when new classes' knowledge is learned as above (Equation~\ref{eq:likelihood}). Based on Equation~(\ref{eq:bayes_log}), the class of the test image $\ve{x}_j$ would be directly predicted as the one with the highest value of $\log p(c \vert \ve{z}_j)$ over all classes learned so far.

The advantages of the the proposed approach over existing continual learning approaches are clear. First, the knowledge of each old class is statistically represented by the set of likelihood functions (Equation~\ref{eq:likelihood}) and compactly stored in the memory. Therefore, old knowledge will not be forgotten over continual learning of new classes. In comparison, old knowledge will be inevitably and gradually forgotten over multiple rounds of continual learning in existing approaches, either due to the changes in feature extractor or due to the reduced number of original images to be stored in the limited memory. Second, the final performance of the proposed approach over multiple rounds of continual learning is not affected by the number of learning rounds and the number of new classes added in each round. In contrast in existing approaches, more rounds of continual learning with smaller number of new classes added each time would often lead to worse classification performance at later round of continual learning. Therefore, the proposed approach is more robust to various continual learning conditions with little forgetting of old knowledge.  

\section{Experimental Evaluations}







\subsection{Experimental setup}

The proposed approach was extensively evaluated on a diverse group of image datasets, including two natural image datasets and two medical skin image datasets. Medical image datasets are normally quite different from natural image datasets in terms of appearance and textures. Each dataset is briefly summarized in the following (also see Table~\ref{tab:dataset}). 

\textbf{CIFAR100}~\cite{krizhevsky2009learning} is a dataset of natural images of daily objects, including various animals, plants, outdoor and indoor scenes, and vehicles etc. It consists of 100 classes, 500 training images and 100 test images for each class. The size of each image is quite small, only $32\times 32$ pixels.

\textbf{CUB200}~\cite{WelinderEtal2010} is a dataset of bird species typically used for fine-grained recognition~\cite{lin2015bilinear}. It contains 11,788 images for 200 categories, approximately 60 images per class. Note that although part locations, binary attributes and bounding boxes are provided, only image-level species labels are used in our classification experiments. For experiments on the CUB200 and the CIFAR100 datasets, the adopted fixed feature extractor is directly from the pre-trained CNN model (e.g., VGG-Net~\cite{simonyan2014very} or ResNet101~\cite{he2016deep}) based on the ImageNet dataset~\cite{deng2009imagenet}, where the last fully connected layer is removed and the remaining part is used for feature extractor. Considering that the pre-trained feature extractor is based on the Imagenet dataset, the TinyImageNet dataset which is a subset of ImageNet and often adopted for continual learning was not used in the experiments here. Also due to the similar reason, the ten common classes (Black footed Albatross, Laysan Albatross, Sooty Albatross, Indigo Bunting, American Goldfinch, Ruby throated Hummingbird, Green Violetear, Blue Jay, Dark eyed Junco and Red breasted Merganser) between CUB200 and ImageNet were removed from CUB200, resulting in 190 classes for continual learning on the CUB200 dataset.

\begin{table}[!tbp]
    \centering
    \captionsetup{justification=raggedright}
    \caption{Datasets for diverse image classification tasks, from natural to medical images, small scale to relatively large scale, and general to fine-grained classifications. Image size varies a lot in Skin40. [120, 500] means that image width and height vary in the range between 120 and 500 pixels.}
    \begin{tabular}{c|c | c | c | c }
    \hline
    Dataset & \#Classes & Training set & Test set & image size \\
    \hline
    CIFAR100 & 100 & 50,000 & 10,000 & $32\times32$\\
    CUB200   & 190 & 5,694  & 5,496  & [120, 500]\\
    Skin7    &  7  & 8010   & 2005   & $600\times450$ \\
    Skin40   &  40 & 2000   & 400    & [260, 1640] \\
    \hline
    \end{tabular}
    \label{tab:dataset}
\end{table}

\textbf{Skin7}~\cite{isic2018dataset} is a skin lesion dataset from the challenge of dermoscopic image classification held by the International Skin Imaging Collaboration (ISIC) in 2018. It consists of 7 disease categories; each image is of size $600\times 450$ pixels. This dataset presents severe class imbalance, with the largest class 60 times larger than the smallest one. 

\textbf{Skin40} is a subset of 193 classes~\cite{skin198sun2016benchmark} of skin disease images collected from the internet. The 40 classes with relatively more number of images (60 images per class) were chosen from the 193 classes to form the Skin40 dataset, while the remaining 153 classes (10 to 40 image per class) were used to train a CNN classifier whose final classification layer was then removed to form the fixed feature extractor in most experiments relevant to the Skin7 and the Skin40 datasets. It is worth noting that there is no overlap between the 153 classes (for training the feature extractor in advance) and the classes in Skin7 and Skin40 (for continual learning evaluation). 
 
During training the feature extractor based on the 153 skin image classes, each image was randomly cropped within the scale range [0.8, 1.0] and then resized to $224\times 224$ pixels, followed by random  horizontal and vertical flipping. The mini-batch stochastic gradient descent (batch size 32) was used to train the feature extractor, with initial learning rate 0.01 and then divided by 10 at the 35th, 70th, and 105th epoch respectively. Weight decay (0.0005) and momentum (0.9) were also applied. The feature extractor was trained for 120 epochs with observed convergence. 

In each experiment, multiple rounds of continual learning were performed, with a few  (e.g., 2, 5, 10) new classes to be learned  at each round. After each round of continual learning, the mean class recall (MCR) over all classes learned so far was calculated. For each experiment, the average and standard deviation of MCR over five runs were reported, where the five orders of classes to be continually learned were fixed and used in the proposed approach and baseline methods. Unless otherwise mentioned, ResNet-101 was used as the backbone for the feature extractor, and the dimension of feature vector $K$ was 2048 and the number ($S$) of Gaussian components in each GMM model was empirically set to 2 based on a small validation set for each dataset.

\subsection{Effectiveness of the generative model}
\begin{figure}[!btp]
    \centering
    \includegraphics[height=0.30\linewidth]{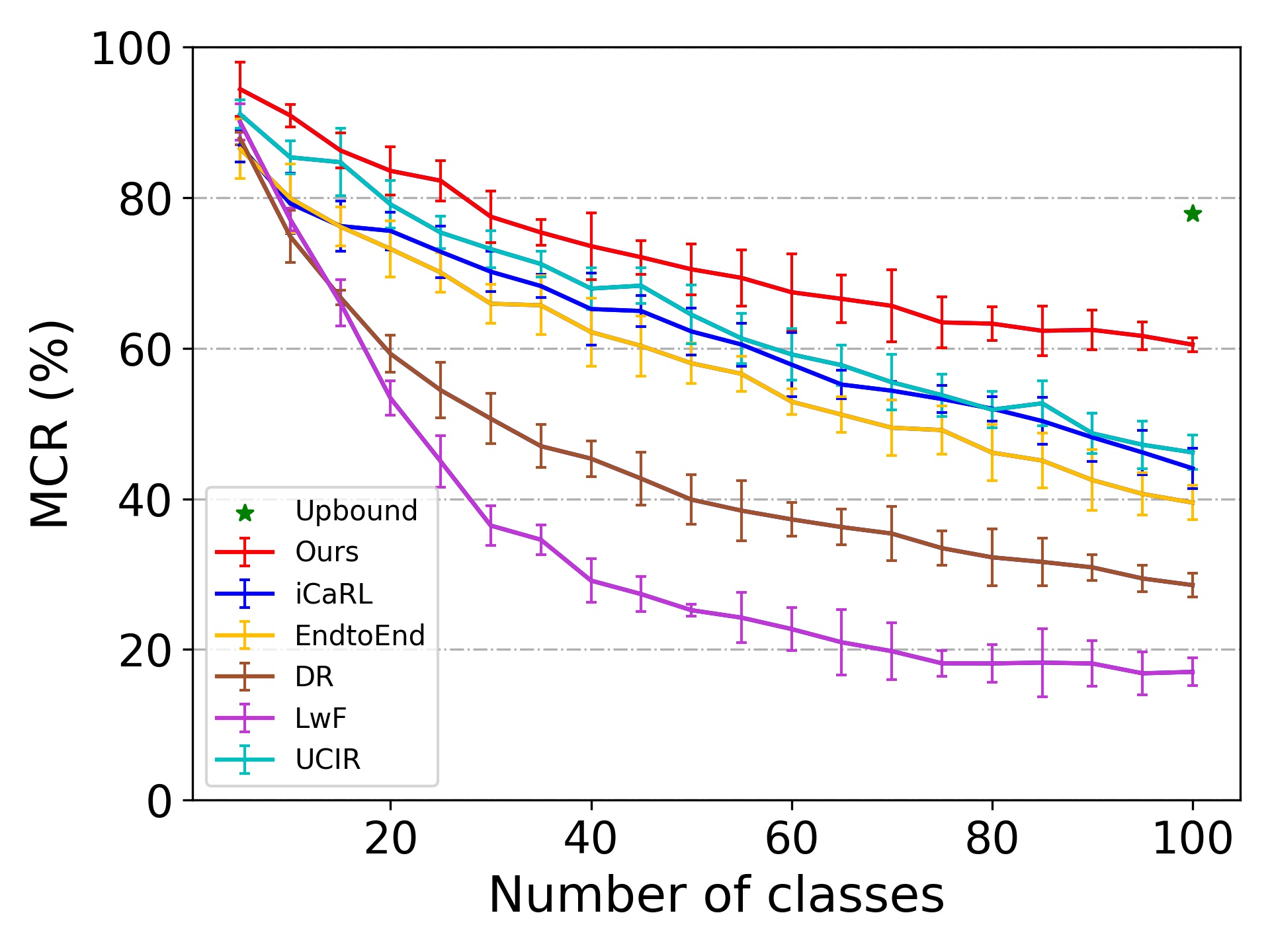}
    \includegraphics[height=0.30\linewidth]{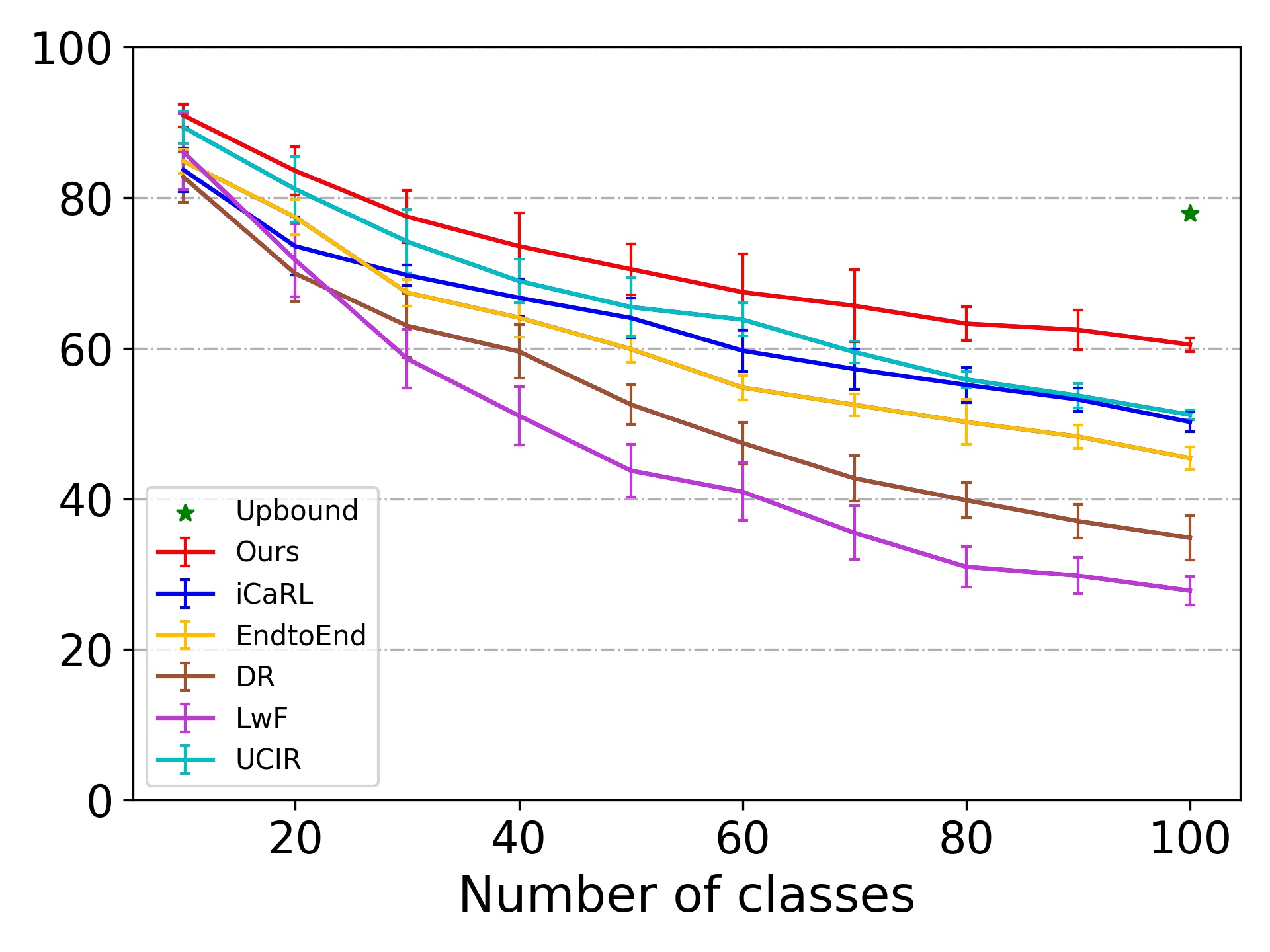}
    \includegraphics[height=0.30\linewidth]{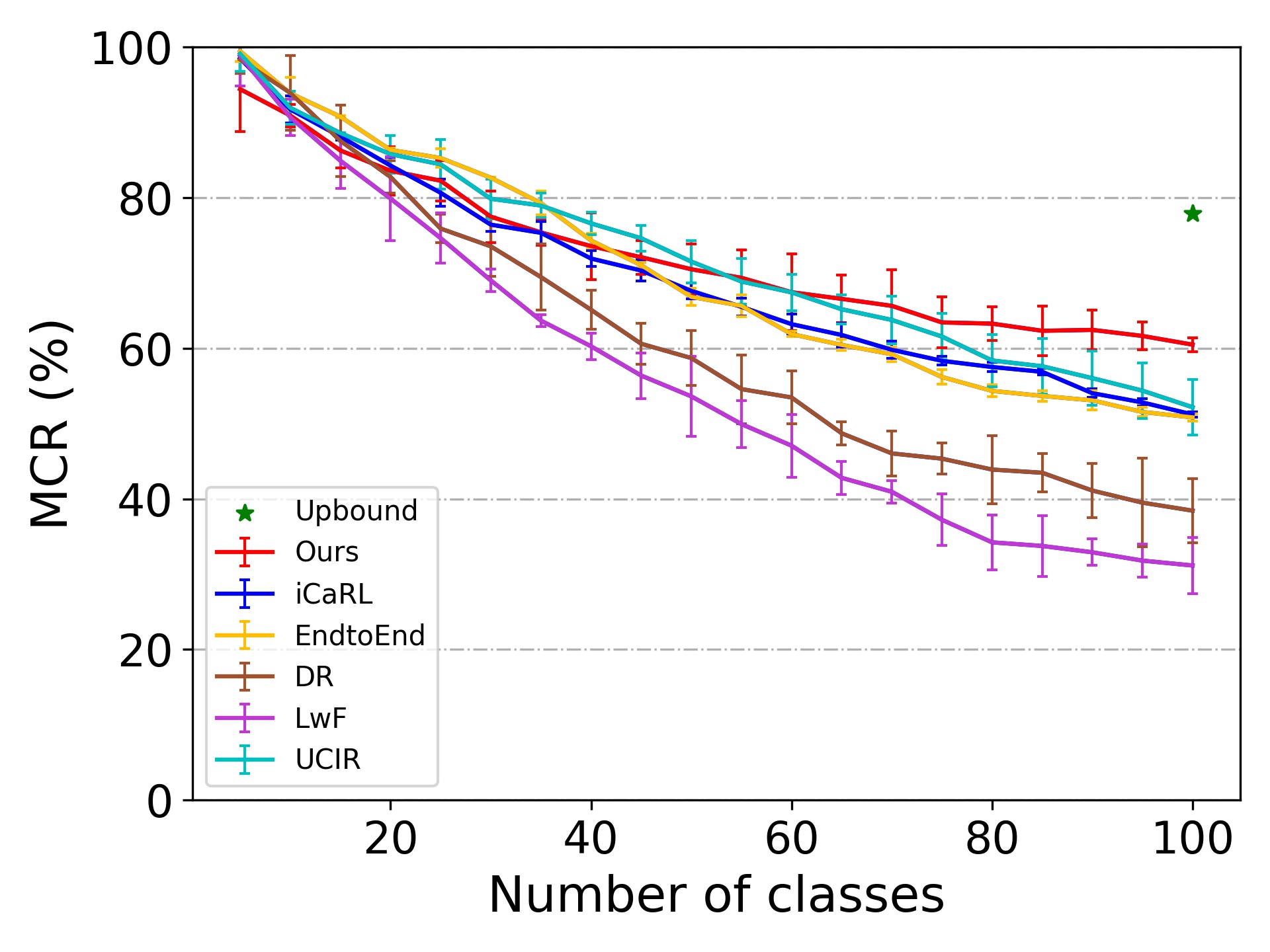}
     \includegraphics[height=0.30\linewidth]{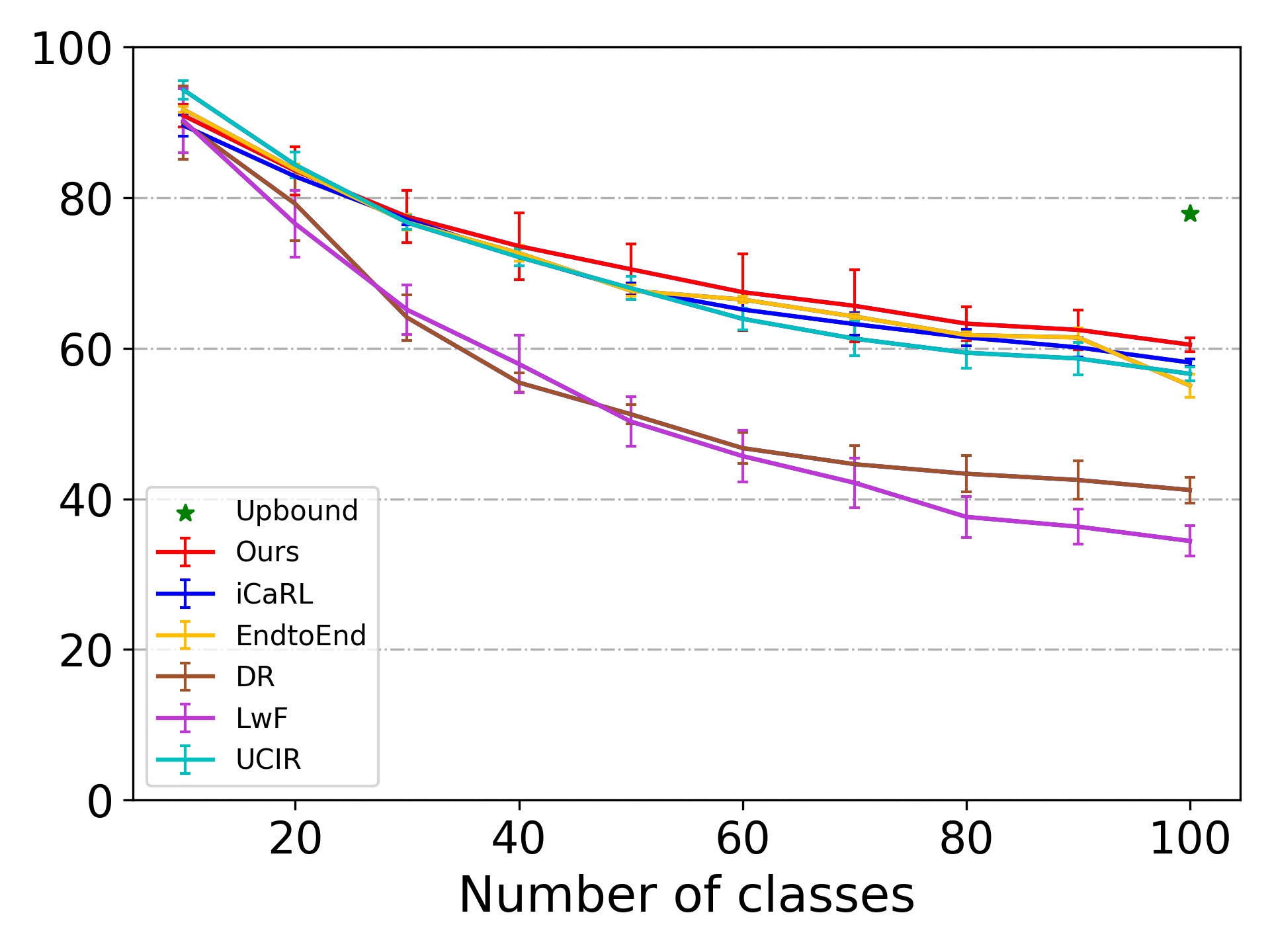}
     \includegraphics[height=0.30\linewidth]{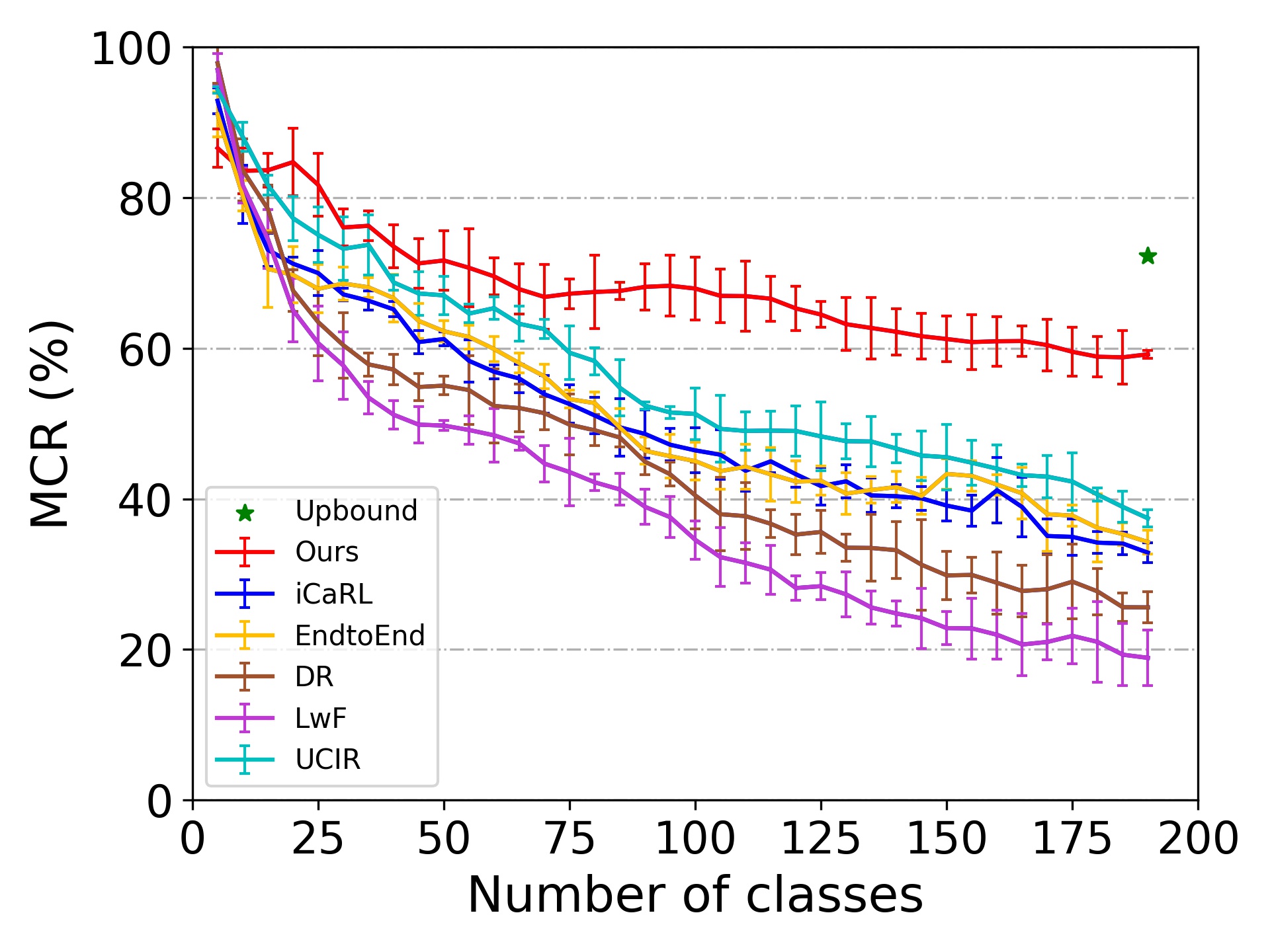}
     \includegraphics[height=0.30\linewidth]{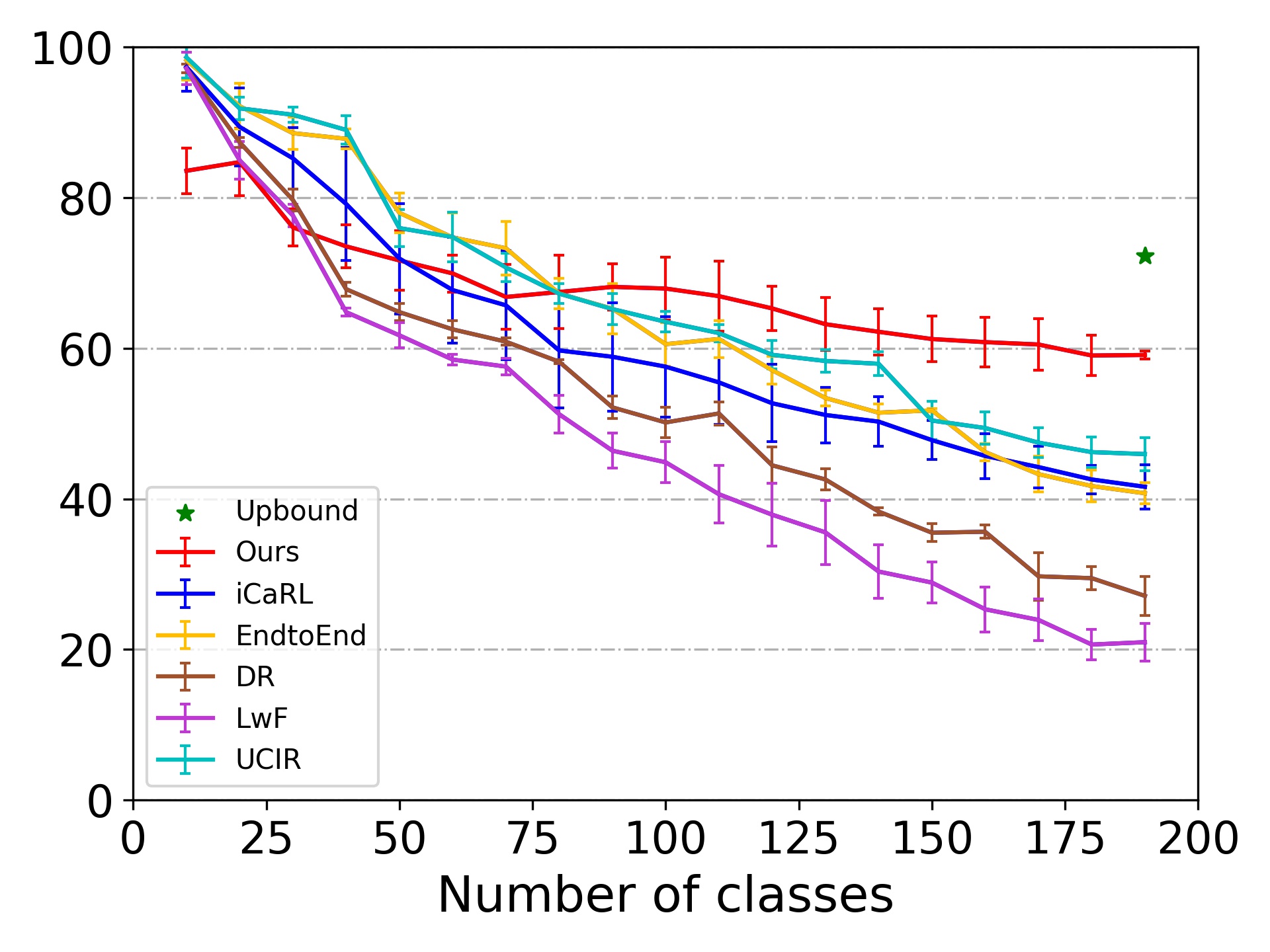}
    \caption{Performance comparison on natural image datasets. First row: learning 5 (Left) and 10 (Right) classes at each round on CIFAR100, with CNN classifiers initially trained from scratch for each baseline method. Second row: learning 5 (Left) and 10 (Right) classes at each round on CIFAR100, with CNN classifiers initially fine-tuned from the same pre-trained feature extractor as that used in the proposed approach. Third row: learning 5 (Left) and 10 (Right) classes at each round on CUB200, with CNN classifiers initially fine-tuned from the same pre-trained feature extractor. X-axis in each sub-figure represents the accumulated number of learned classes in the corresponding continual learning task.}
    \label{fig:effective_natural}
\end{figure}

\begin{figure*}[!btp]
    \centering
    \includegraphics[height=0.2\linewidth]{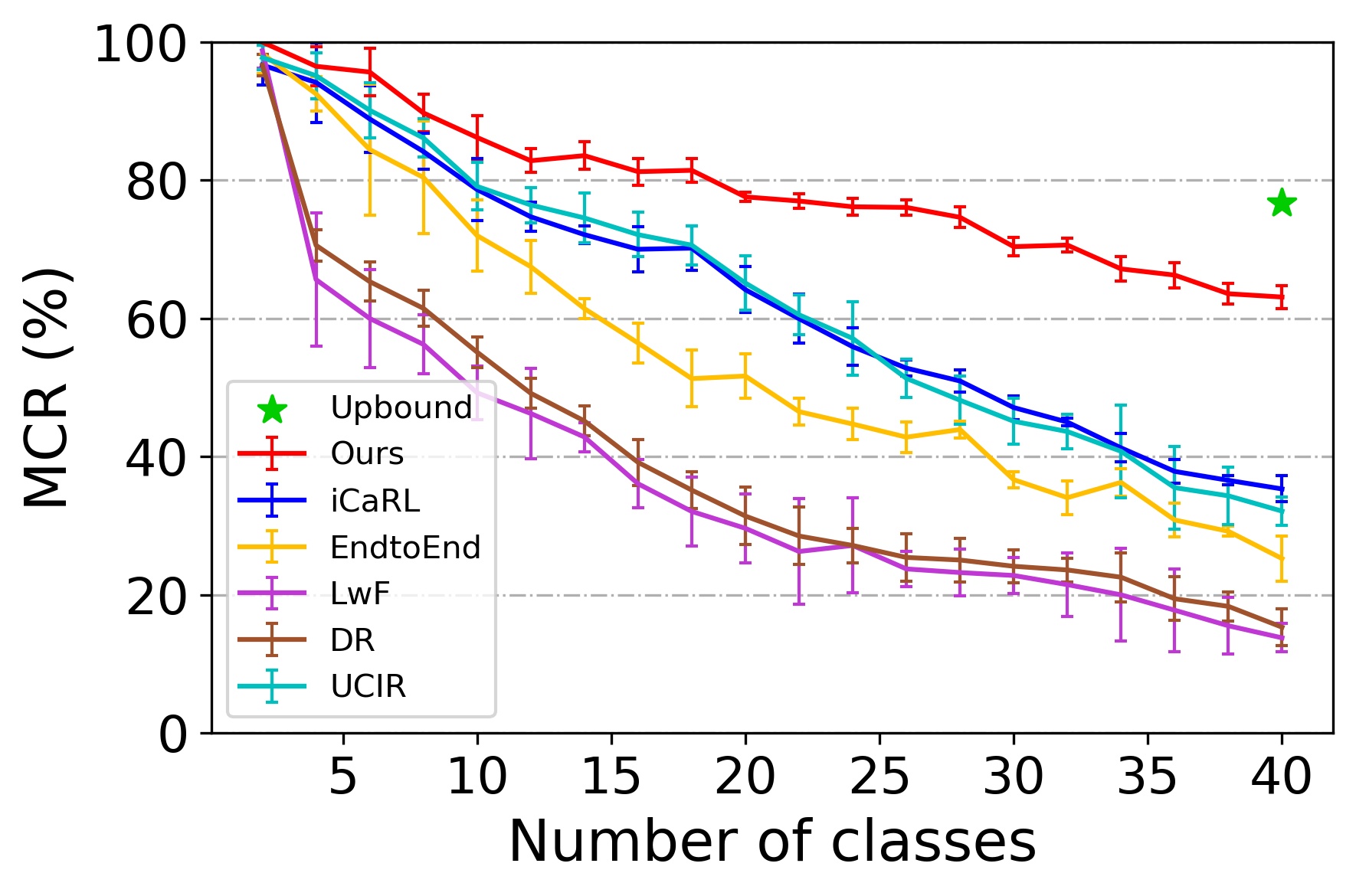}
    \includegraphics[height=0.2\linewidth]{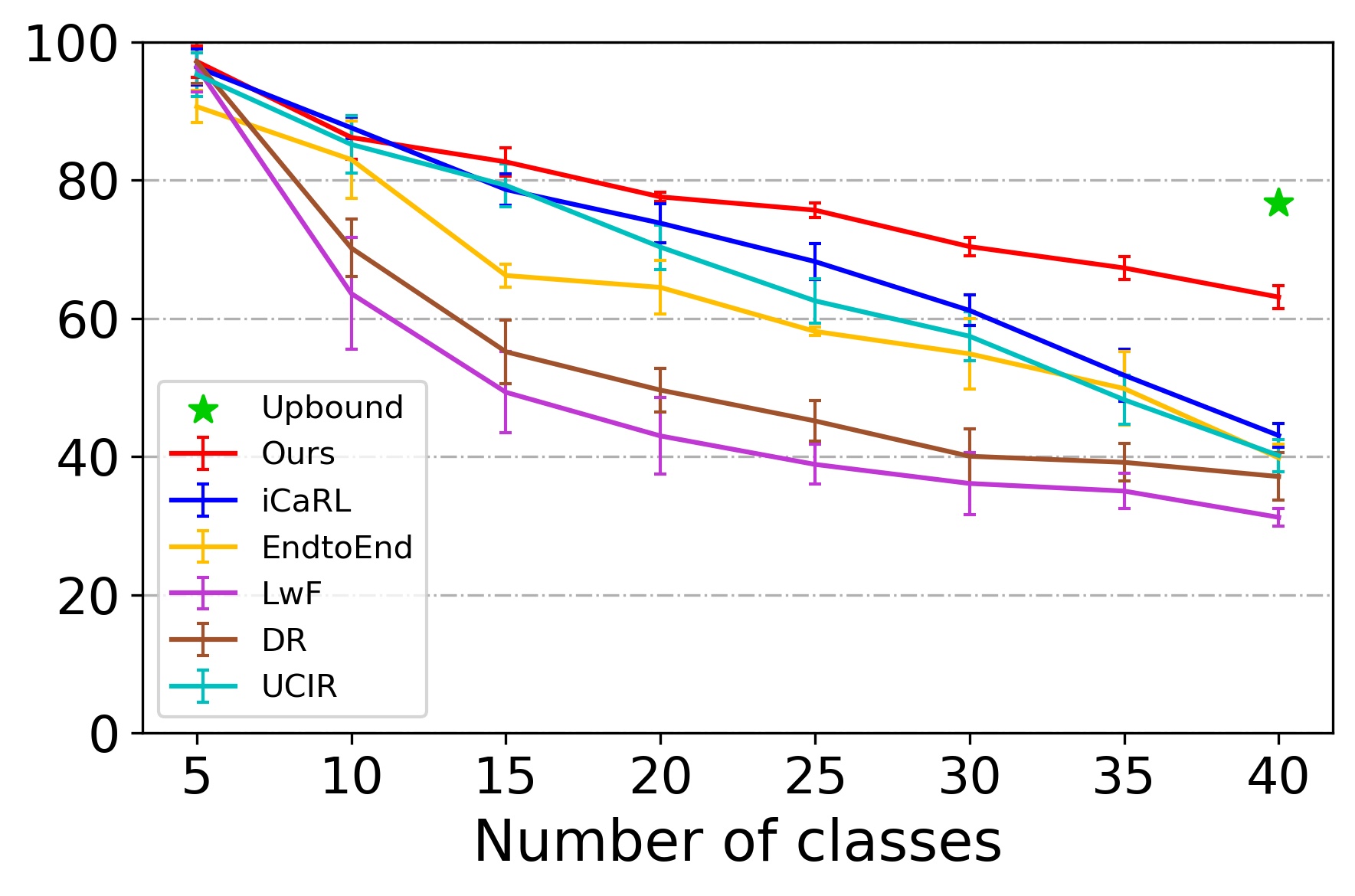}
    \includegraphics[height=0.2\linewidth]{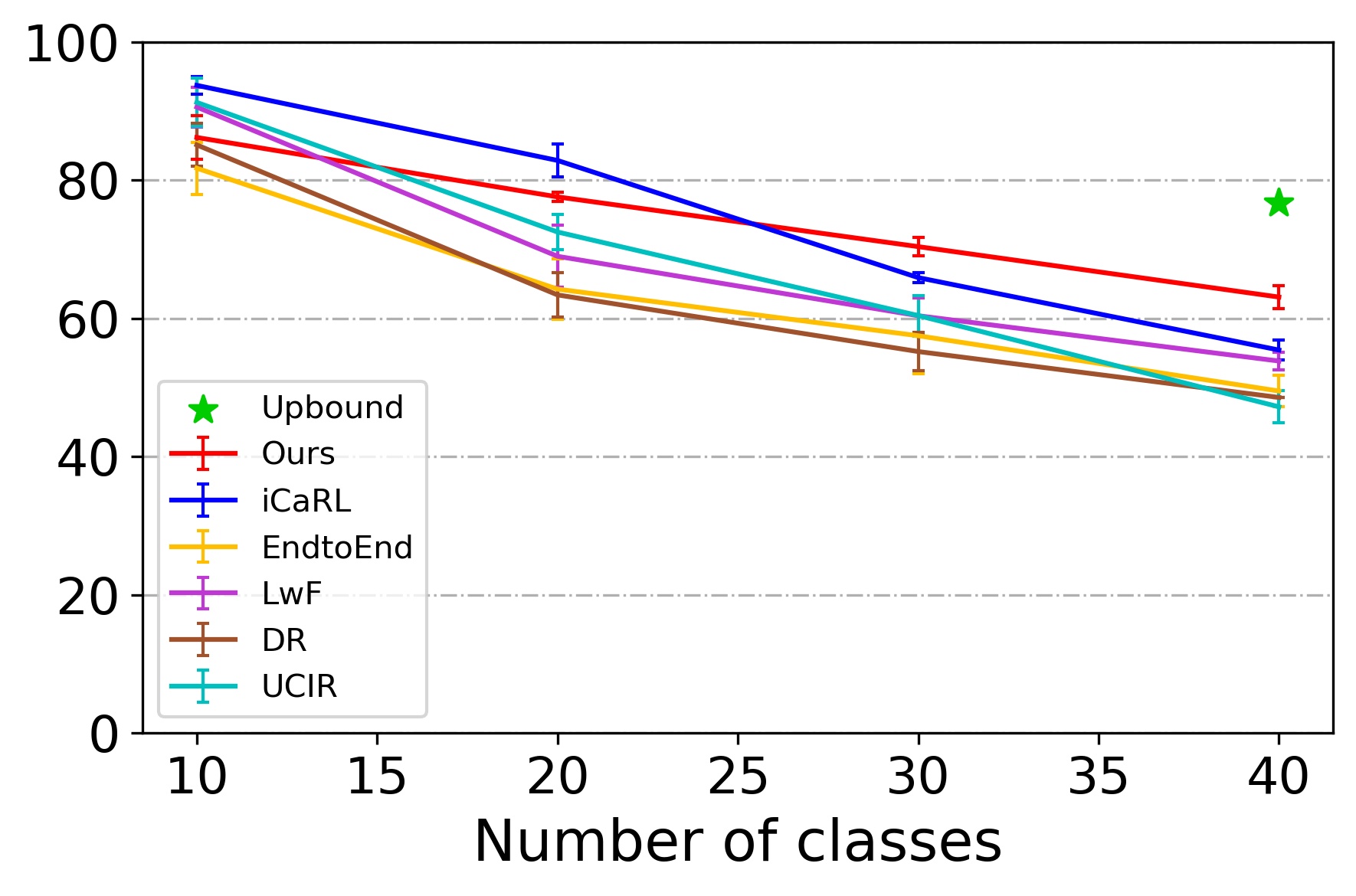}
    \includegraphics[height=0.2\linewidth]{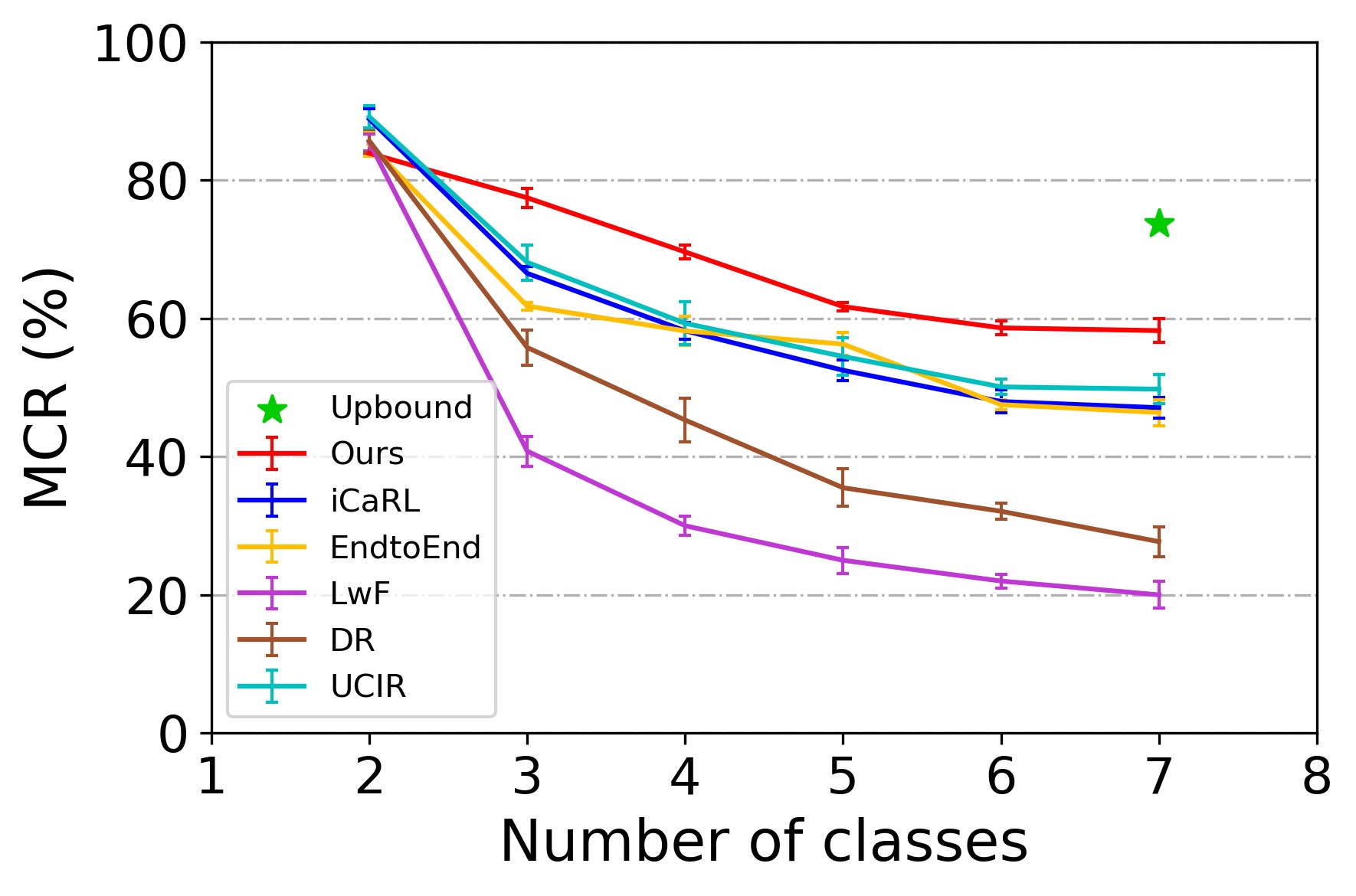}
    \includegraphics[height=0.2\linewidth]{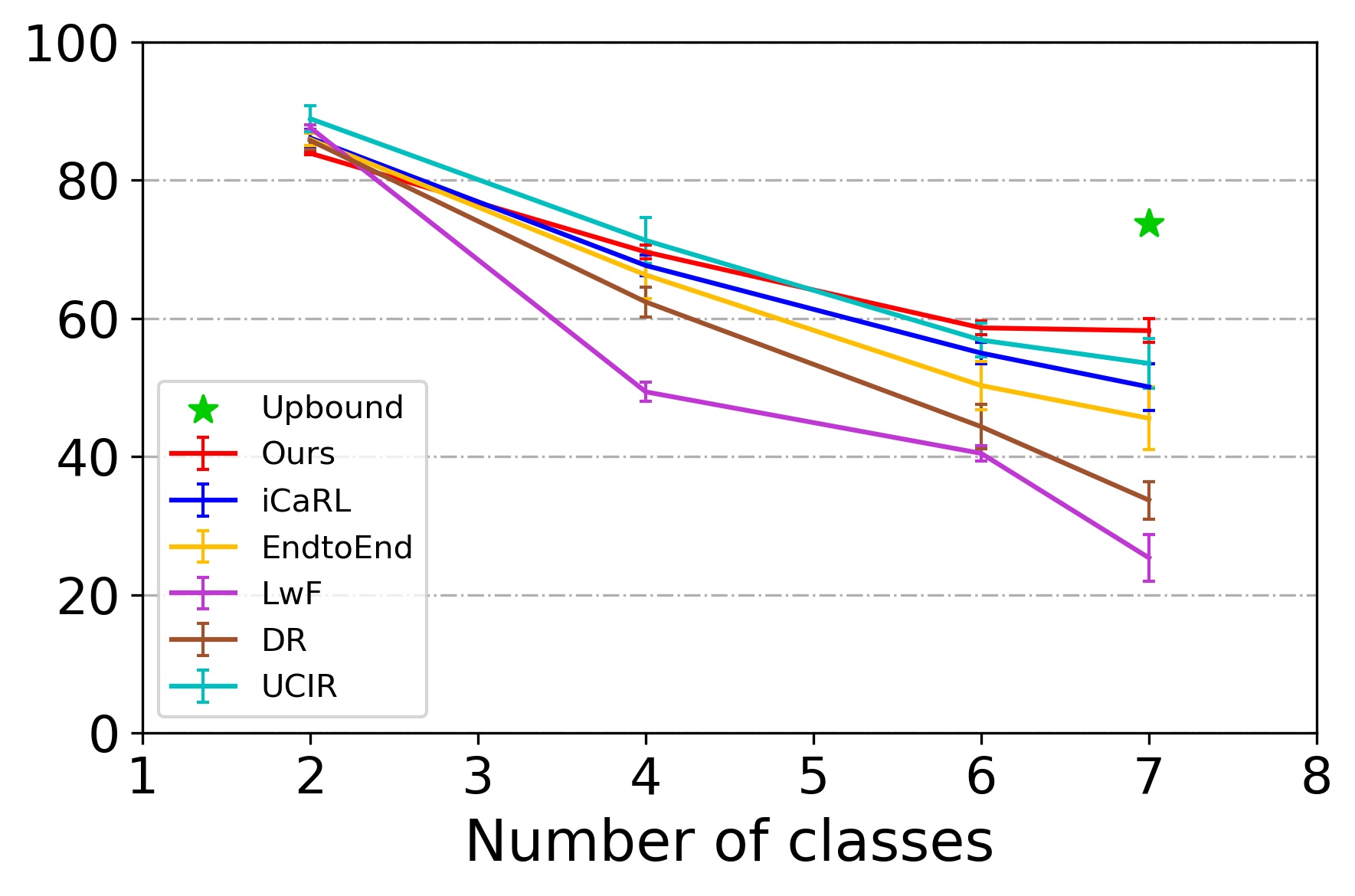}
    \includegraphics[height=0.2\linewidth]{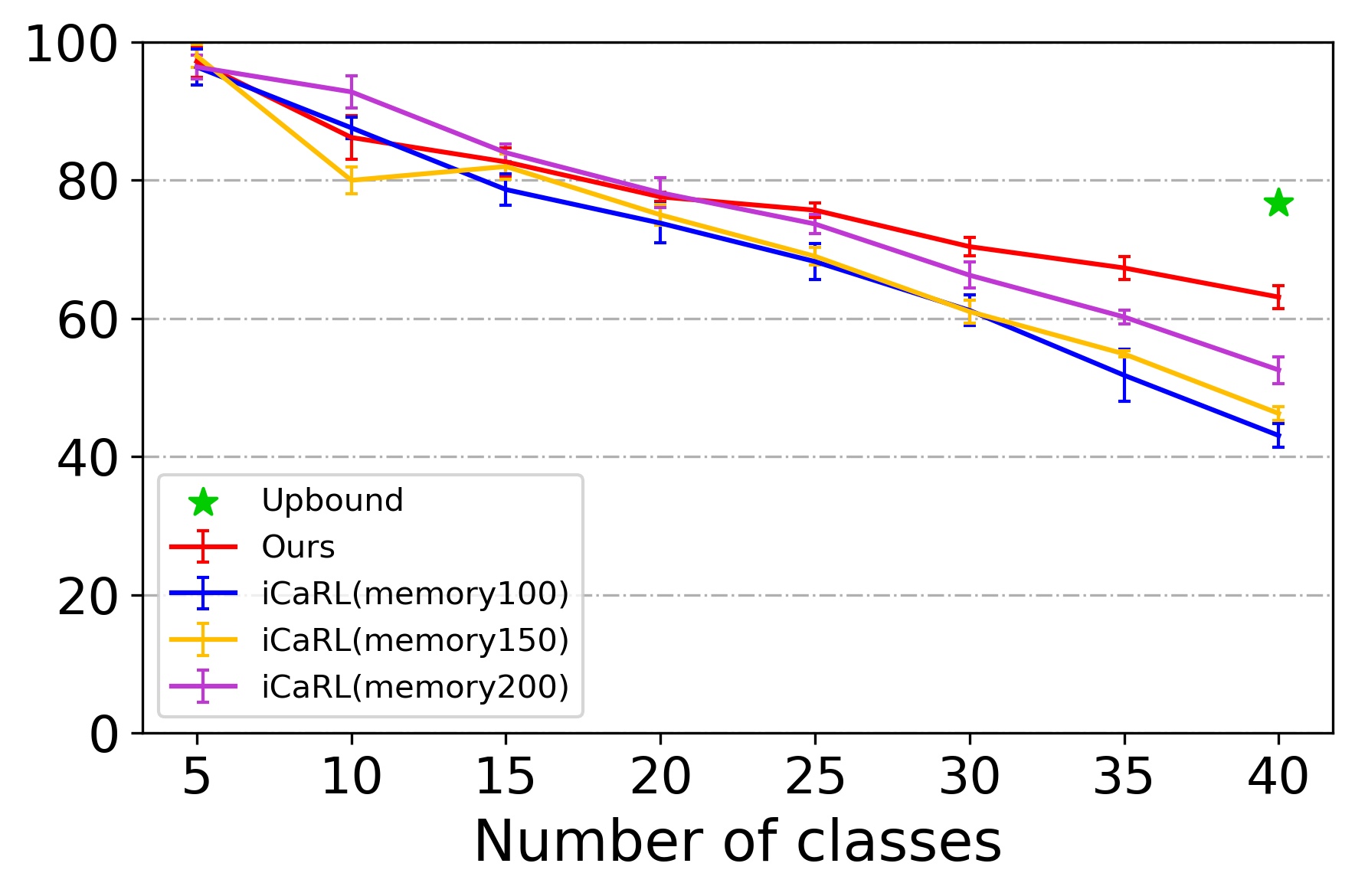}
    \caption{Performance comparison on medical image datasets. First row (from left): learning 2, 5, 10 classes at each round on Skin40. Second row (from left): learning 1 and 2 classes at each round on Skin7, and {comparison with iCaRL with varying memory size on Skin40}.  X-axis in each sub-figure represents the accumulated number of learned classes in the corresponding continual learning task.}
    \label{fig:effective_skin}
\end{figure*}

This section evaluates the effectiveness of the proposed approach by comparing with state-of-the-art strong baselines, including iCaRL~\cite{rebuffi2017icarl}, End-to-End Incremental Learning (End2End)~\cite{castro2018end}, learning a Unified Classifier Incrementally via Rebalancing (UCIR)~\cite{hou2019learning}, Distillation and Retrospection (DR)~\cite{hou2018lifelong}, and Learning without Forgetting (LwF)~\cite{li2017learning}. The suggested hyper-parameter settings in the original work were adopted unless otherwise mentioned. 
In each round of continual learning, for the iCaRL, End2End, DR, and UCIR which need certain number of old data, the number of images stored (i.e., memory size) for all old classes is respectively 2000 on CIFAR100, 400 on CUB200, 50 on Skin7, and 100 Skin40. The memory size was chosen such that stored number of images for each class was only a small portion of the original training images at the last round of continual learning. For each experiment, an upper-bound result was also reported (e.g., Figure~\ref{fig:effective_natural} and Figure~\ref{fig:effective_skin}, {green star}) by training a non-continual classifier with all classes of training data.

All the baselines were previously evaluated by initially training a CNN classifier from scratch before starting continual learning. Therefore, the proposed approach  was firstly compared to the baselines where each initial CNN classifier for each baseline was trained from scratch. In this case, Figure~\ref{fig:effective_natural} (First row) clearly shows that the proposed approach outperforms all the strong baselines. Compared to the strongest baseline, the absolute improvement by the proposed approach is respectively 14.3\% (First row, Left) and 9.3\% (First row, Right) at the last round of continual learning when respectively learning 5 and 10 new classes at each round. However, the comparison could be considered unfair because the proposed approach used a pre-trained feature extractor while the baselines did not. In consideration of this point, the proposed approach was also compared with the baselines where each initial CNN classifier at the first learning round was fine-tuned from the same pre-trained feature extractor for each baseline method. In this case, Figure~\ref{fig:effective_natural} (Second row) demonstrates that although some strong baselines have slightly better performance at the first several rounds of continual learning, the performance of all baselines decreases faster than the proposed approach, and is outperformed by the proposed approach at the last several rounds of continual learning. The more rounds of continual learning, the larger final gap between the proposed approach and the strongest baseline at the last round of continual learning. This is further confirmed on the CUB200 dataset where more rounds of continual learning happened compared to on  CIFAR100 with the same settings. As shown in Figure~\ref{fig:effective_natural} (Third row), the classification performance of the proposed approach decreases much slower than that of all the baselines, and the proposed approach quickly outperforms all the strong baselines after a few rounds of continual learning, although the same pre-trained feature extractor was initially used to fine-tune each CNN classifier in each baseline. It can be consistently observed that the gap between the strongest baseline and the proposed approach becomes larger and larger with more rounds of continual learning. 

As expected, Figure~\ref{fig:effective_natural} also tells us the final-round performance of the proposed approach is not affected by the number of new classes to be learned at each round. The performance of the proposed approach is around 60\% in MCR at the last round of continual learning on both the CIFAR100 and the CUB200 datasets, no matter how many rounds of continual learning are performed and how many new classes are learned at each round. In comparison, the final performance of each baseline becomes worse with more rounds of continual learning (correspondingly with smaller number of new classes to be learned at each round). These results clearly support that the proposed Bayesian generative model is effective in reducing the catastrophic forgetting of old knowledge, probably because knowledge of old classes were kept unchanged in the form of statistical distribution over continual learning.

The proposed approach works effectively not only on the natural image datasets, but also on medical datasets. Figure~\ref{fig:effective_skin} shows that, with certain number of new classes to be continually learned at each round on both Skin40 and Skin7 datasets, the proposed approach always performs better than all the strong baselines particularly at later round of continual learning, although the same pre-trained feature extractor was used to initially fine-tune the CNN classifier for each baseline method (which is the default setting in the following sections). Even with more images of old classes stored for the representative strong baseline iCaRL, the proposed approach still performs better (Figure~\ref{fig:effective_skin}, second row, last), again supporting that the proposed approach is more effective in keeping old knowledge from forgetting.

\begin{figure}[!btp]
    \centering
    \includegraphics[height=0.30\linewidth]{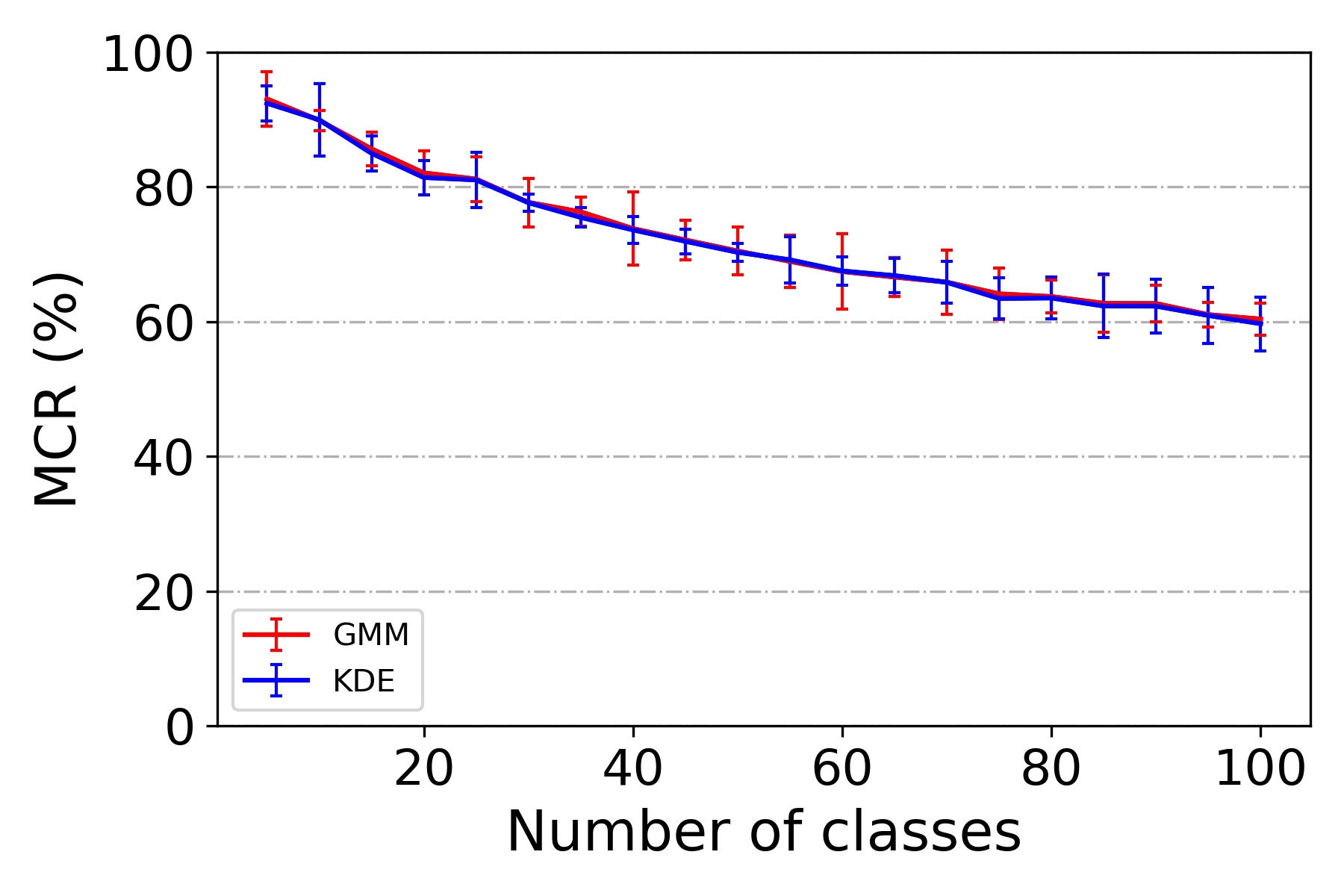}
    \includegraphics[height=0.30\linewidth]{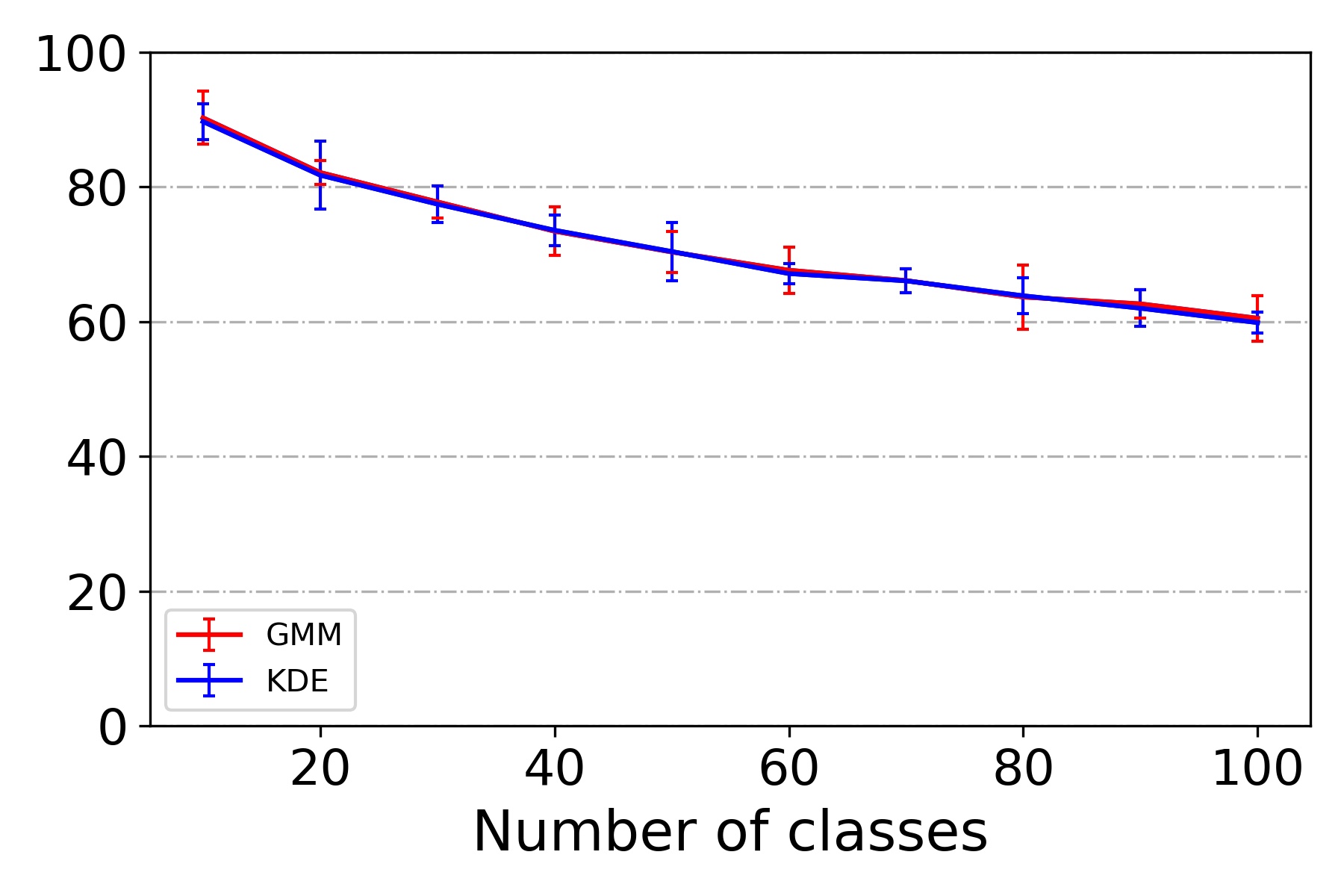}
    \includegraphics[height=0.30\linewidth]{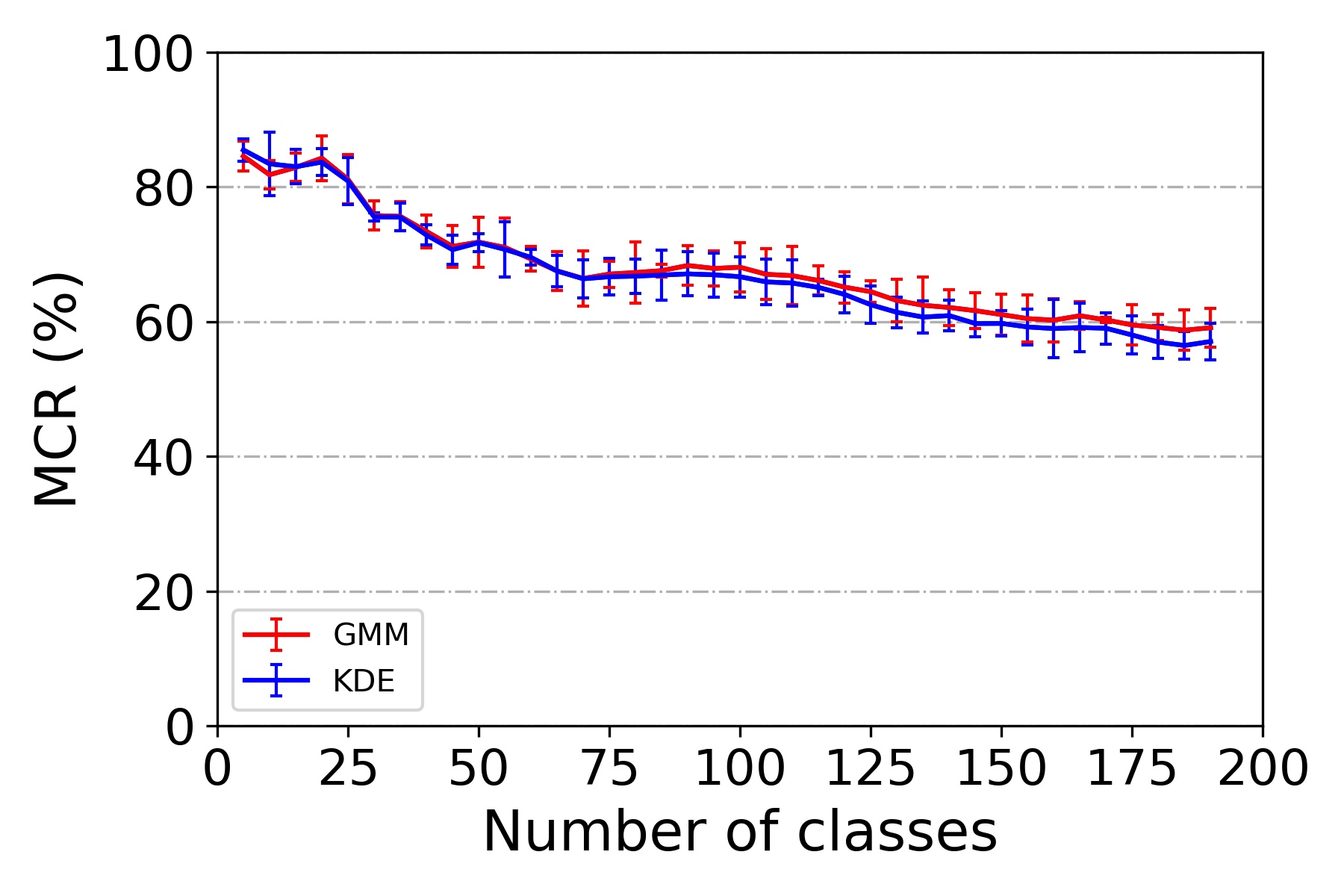}
    \includegraphics[height=0.30\linewidth]{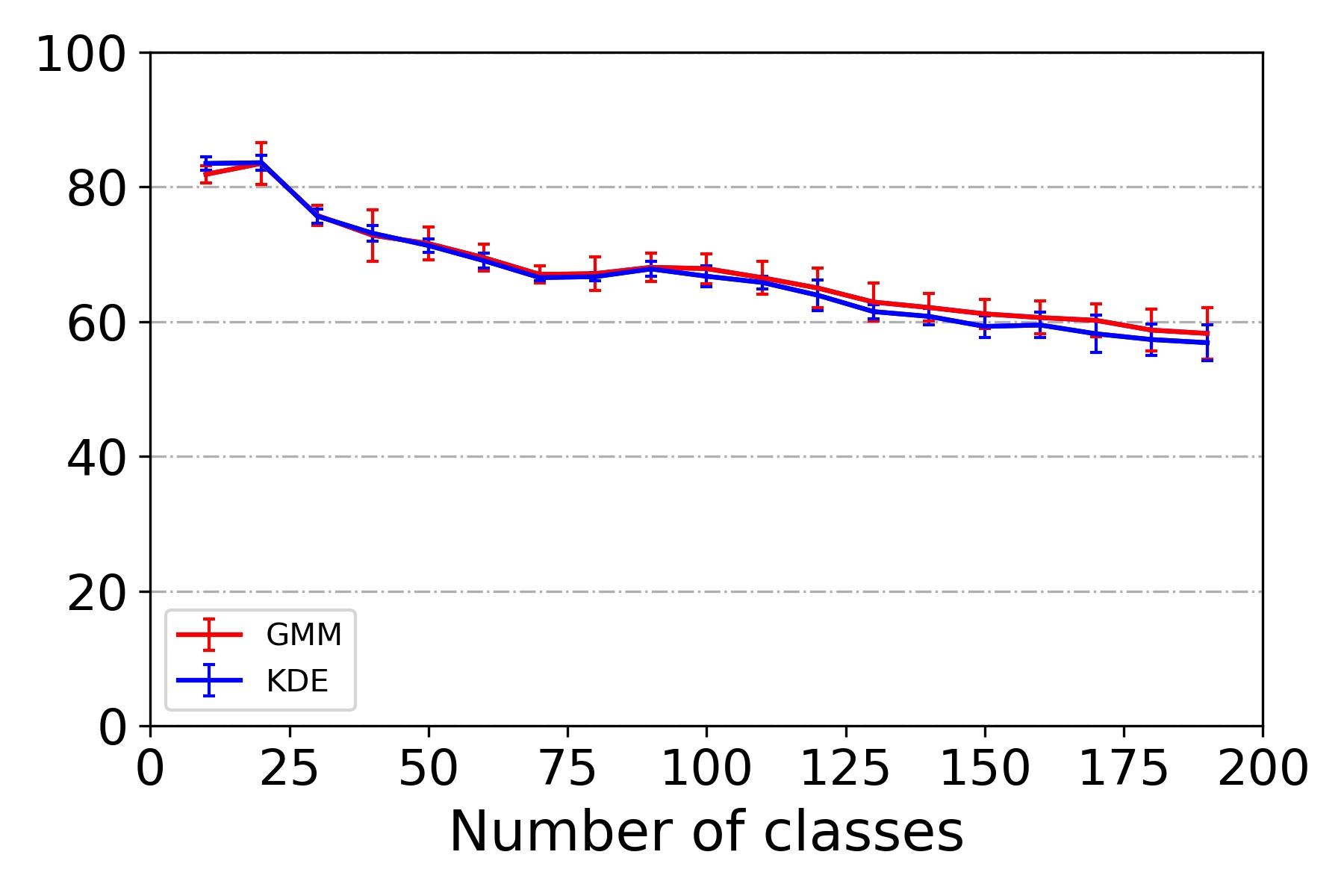}
    \caption{Performance comparison between GMM and KDE models for old knowledge representation respectively on CIFAR100 (first row) and CUB200 (second row), when learning 5 (first column) or 10 (second column) new classes at each round.}
    \label{fig:kde}
\end{figure}

\subsection{Generalizability and robustness of the generative model}

The proposed approach is a general framework and therefore can employ different feature extractor backbones or use different ways to represent and store old knowledge in specific applications. As Table~\ref{tab:backbone} shows, the proposed approach performs consistently better than strong baselines on all the four datasets with different feature extractor backbones (Vgg19, ResNet18, ResNet34, ResNet101), supporting that the proposed approach is not limited to specific feature extractor structures. 
Also, the proposed approach is not limited to the specific Gaussian Mixture Model (GMM) representation for old knowledge. For example, besides the parametric GMM model, the well-known nonparametric Kernel Density Estimation (KDE) was also used to approximate the statistical distribution of each feature output, where the kernel width is empirically determined based on a small validation set from each dataset.
Figure~\ref{fig:kde} shows that the proposed approach based on KDE works equivalently well compared to that based on GMM for representation of old knowledge. Because the proposed approach is not limited to specific ways to represent old knowledge, potentially more effective representation of old knowledge would increase the performance of continual learning by the proposed approach. This remains to be explored in future work. 
  

\begin{table*}[!tbp]
\renewcommand\arraystretch{1.2}
\caption{Performance on various feature extractor backbones. Results after last round of continual learning were reported, with 1 (Skin7), 5 (Skin40), 5 (CIFAR100), and 5 (CUB200) new classes per round.}
\resizebox{\linewidth}{!}{
    \begin{tabular}{@{}crrrrcrrrrcrrrrcrrrr@{}}
        \toprule
        \multirow{2}{*}{Dataset} & \multicolumn{4}{c}{VGG19}    &  
                                 & \multicolumn{4}{c}{ResNet18} & 
                                 & \multicolumn{4}{c}{ResNet34} &
                                 & \multicolumn{4}{c}{ResNet101}\\
        \cmidrule{2-5} \cmidrule{7-10} \cmidrule{12-15} \cmidrule{17-20}
        &LwF  &iCaRL  &IR  &Ours  &
        &LwF  &iCaRL  &IR  &Ours  &
        &LwF  &iCaRL  &IR  &Ours  &
        &LwF  &iCaRL  &IR  &Ours \\
        \midrule
        CIFAR100 & 10.7  & 35.3  & 39.4 & \textbf{48.9}  &
                 & 12.2  & 38.4  & 43.7 & \textbf{57.2}  &
                 & 12.7  & 39.5  & 44.5 & \textbf{61.8}  &
                 & 12.9  & 40.2  & 46.2 & \textbf{60.5} \\
                 
        CUB200   & 11.3  & 28.1  & 26.6 & \textbf{46.3}  &
                 & 14.2  & 30.1  & 32.2 & \textbf{54.9}  &
                 & 14.4  & 31.2  & 33.7 & \textbf{56.5}  &
                 & 15.2  & 33.6  & 35.2 & \textbf{59.2} \\
                 
        Skin7    & 18.9  & 39.7  & 38.3 & \textbf{46.5} &
                 & 19.8  & 44.3  & 46.2 & \textbf{55.6} &
                 & 20.1  & 46.9  & 48.3 & \textbf{56.8} &
                 & 21.0  & 48.5  & 50.0 & \textbf{58.4}\\ 

        Skin40   & 27.4  & 33.6  & 32.5 & \textbf{52.8} &
                 & 30.4  & 41.8  & 37.1 & \textbf{61.9} &
                 & 31.1  & 42.3  & 39.5 & \textbf{62.8} &
                 & 31.2  & 43.1  & 40.2 & \textbf{63.1}\\ 
        \bottomrule
    \end{tabular}
}
\label{tab:backbone}
\end{table*}

\begin{figure}[!tpb]
    \centering
    \includegraphics[height=0.3\linewidth]{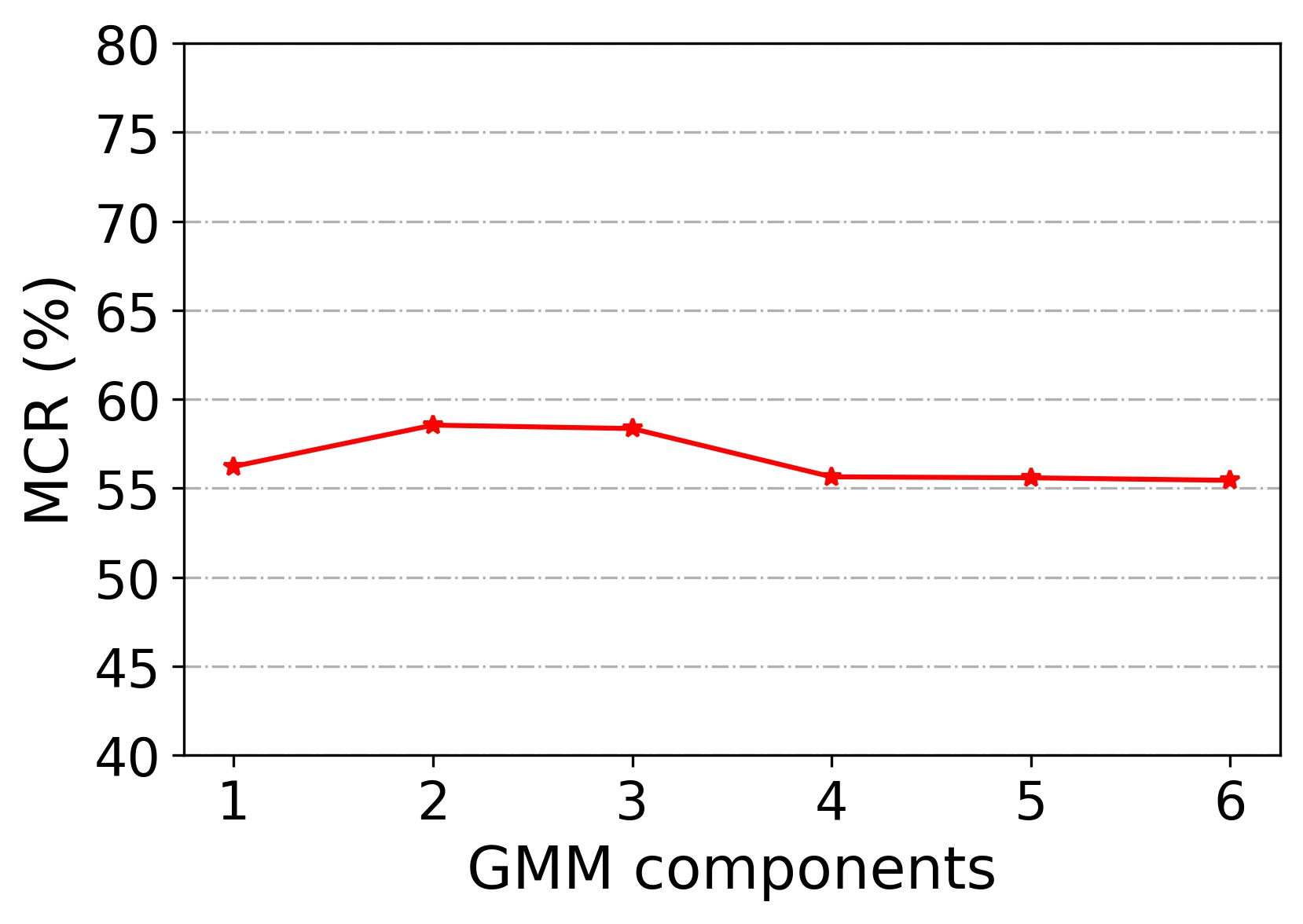}
    \includegraphics[height=0.3\linewidth]{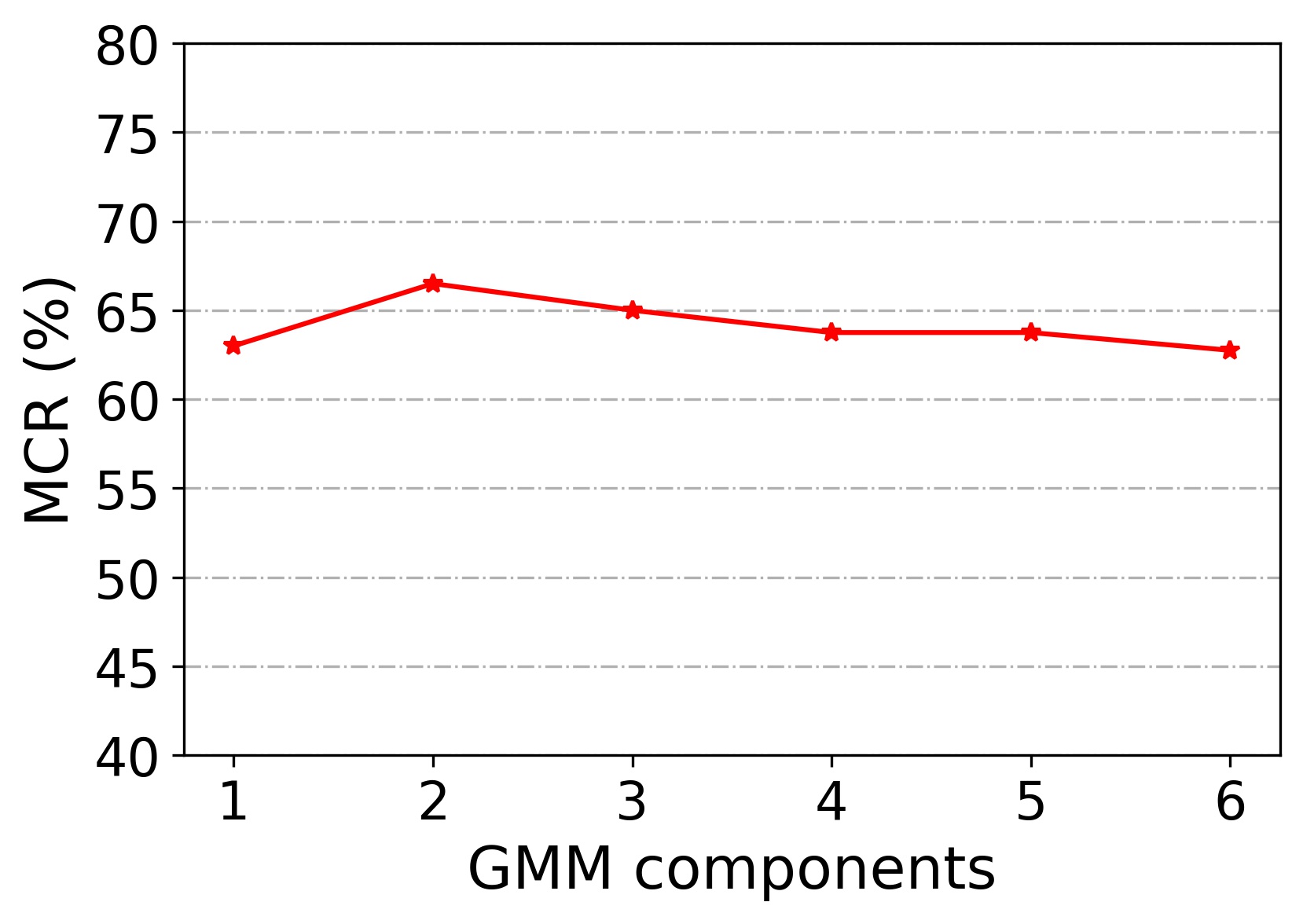}
    \includegraphics[height=0.3\linewidth]{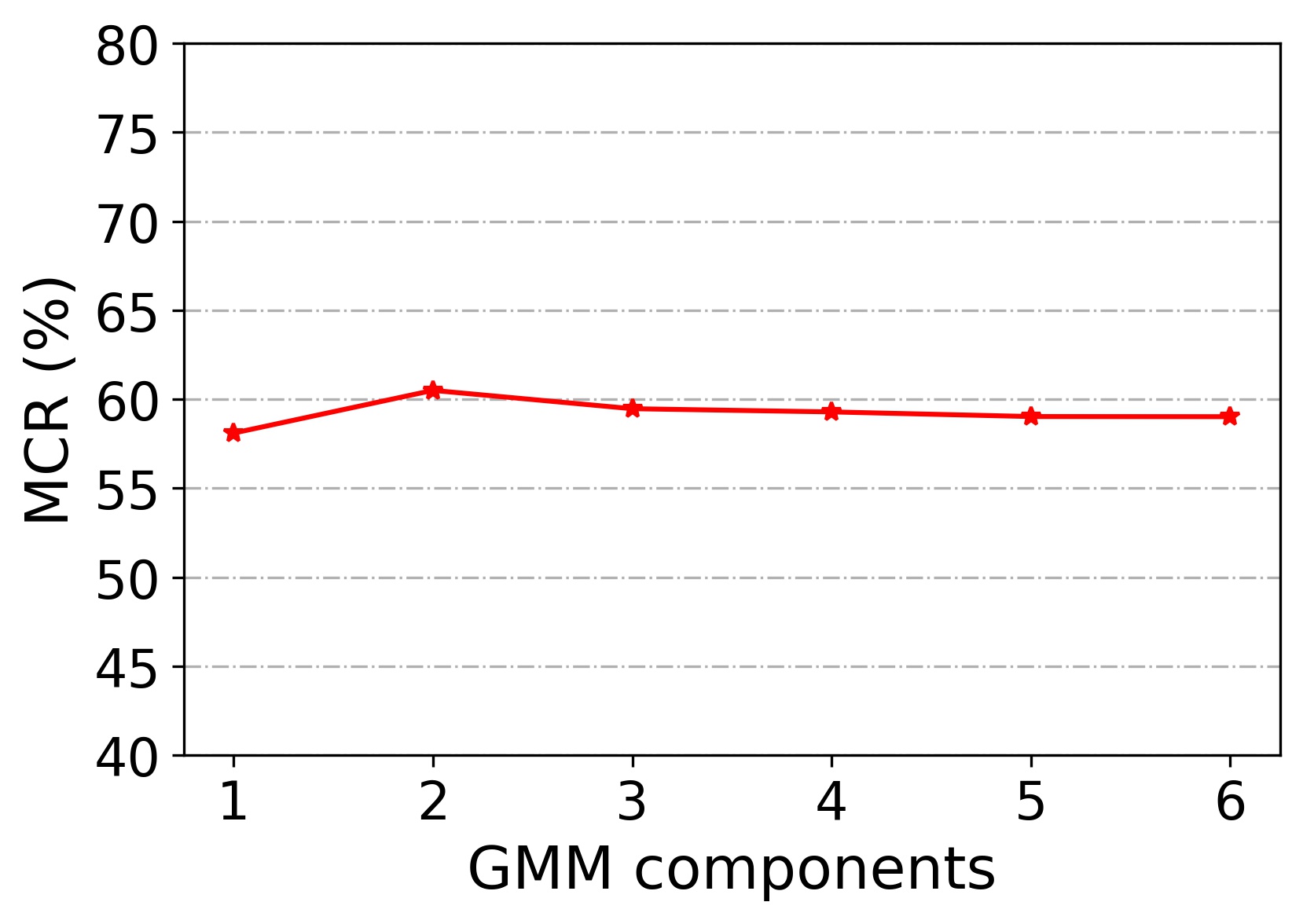}
    \includegraphics[height=0.3\linewidth]{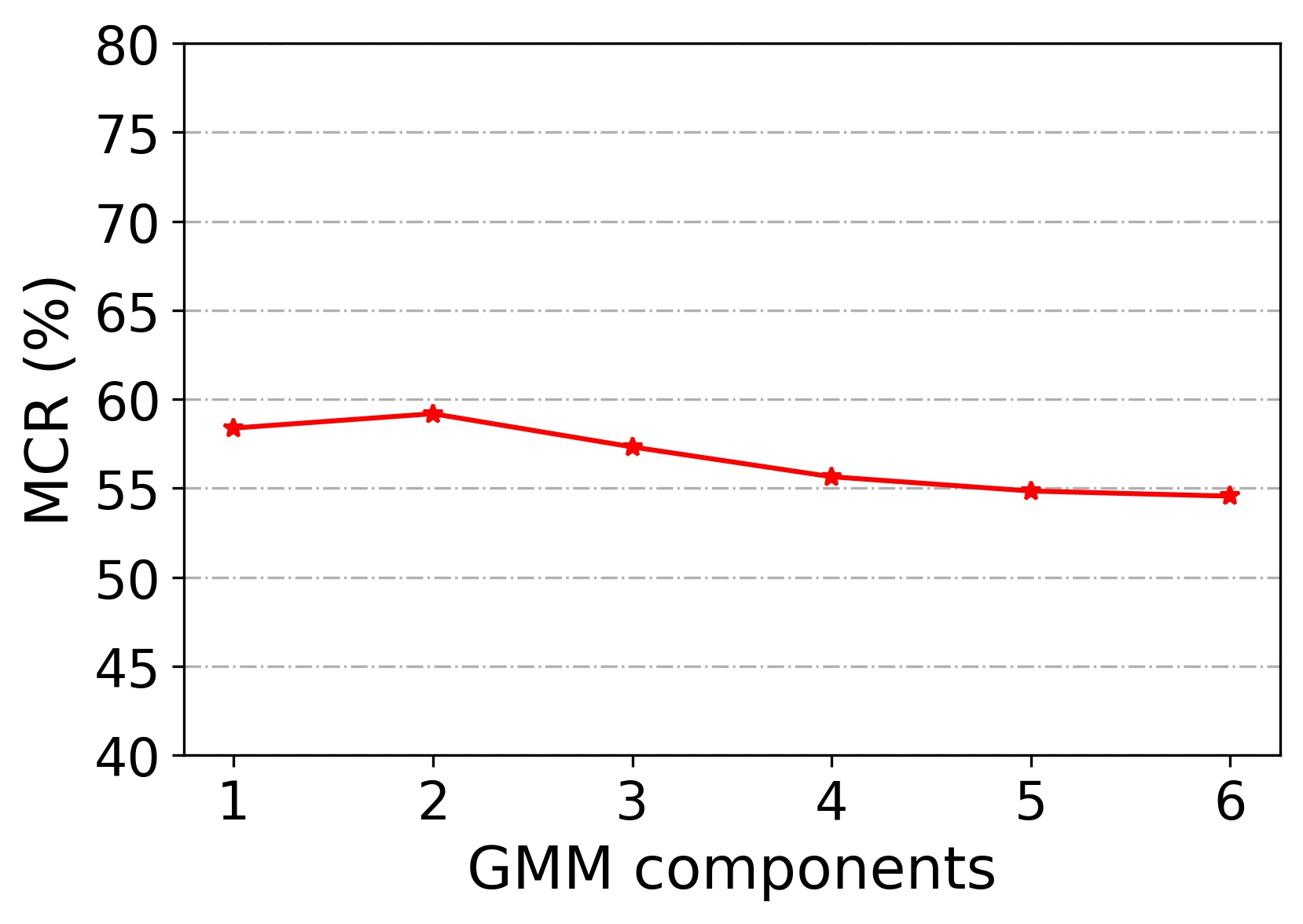}
    \caption{Stable performance with varying GMM components on Skin7 (first row, left), Skin40 (first row, right), CIFAR100 (second row, left) and CUB200 (second row, right). At each round of continual learning, 1, 5, 10, and 10 new classes were continually learned respectively on Skin7, Skin40, CIFAR100, and CUB200 datasets.}
\label{fig:robust_gmm}
\end{figure}

\begin{figure}[!tpb]
    \centering
    \includegraphics[width=0.60\linewidth]{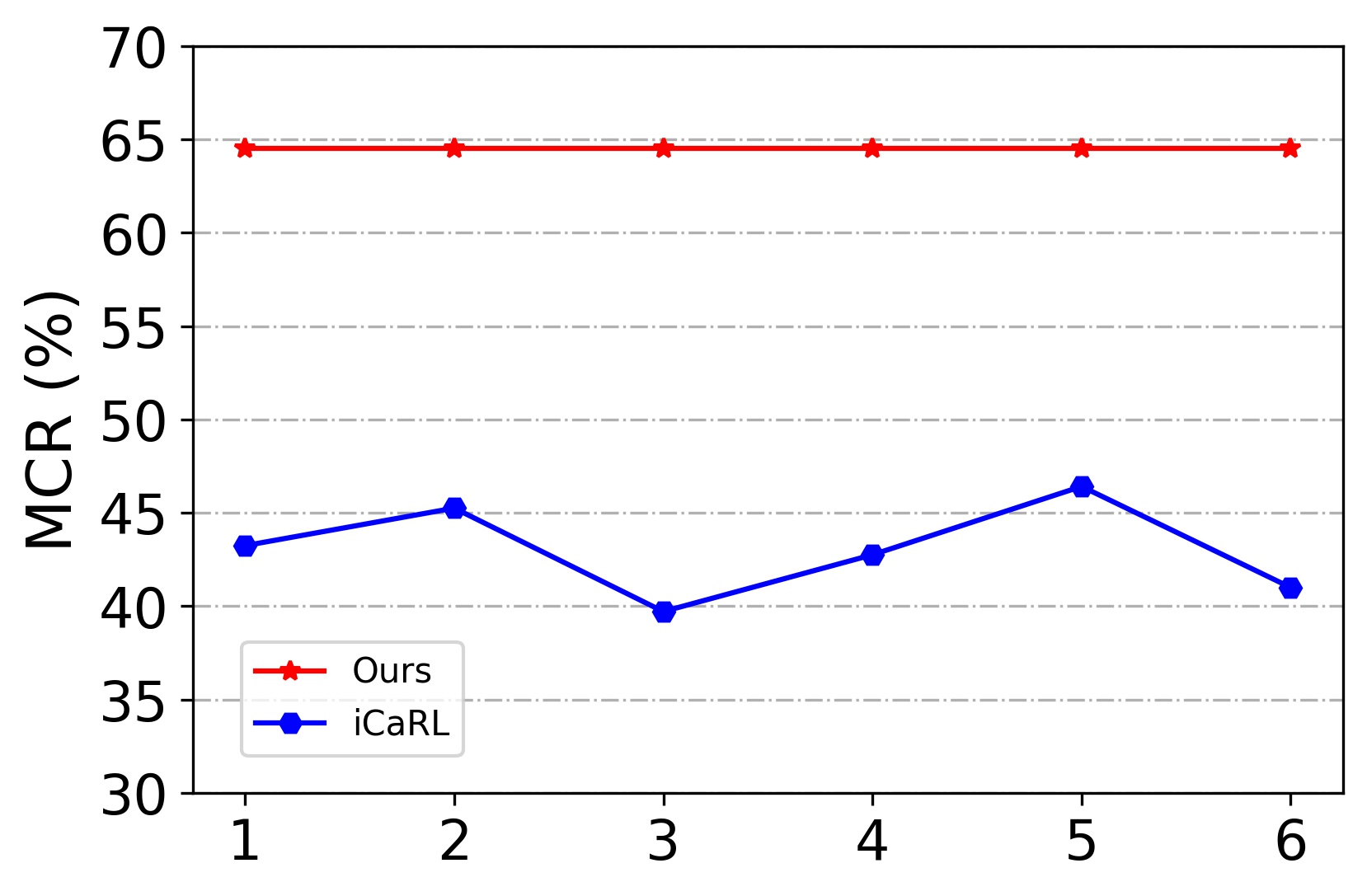}
    \caption{Final-round learning performance with different class orders during continual learning on Skin40. Five new classes were continually learned at each round. X-axis indices represent different sequences of class orders.}
    \label{fig:robust_order}
\end{figure}

To evaluate the robustness of the generative model, the GMM with varying number of Gaussian components and different orders of classes to be continually learned were tried during continual learning. Figure~\ref{fig:robust_gmm} clearly shows that the generative model works stably with different number of Gaussian components in the GMM on all four datasets, although GMM with two components works slightly better than GMM with fewer or more components. Note that the proposed approach still outperforms all the strong baselines when the number of GMM components is larger than two. 
In addition, with six different orders of classes to be continually learned, the performance of the proposed approach does not change at the last round of continual learning, while the performance of the representative iCaRL baseline method clearly varies  with different class orders (Figure~\ref{fig:robust_order}). This is because  knowledge of each previously learned old class is compactly stored and not changed throughout the whole process of continual learning by the proposed approach. In comparison, almost all the strong baseline methods inevitably update the feature extractor during continual learning, which would then change the representation of each stored old data and further change the representation of old knowledge, differently with different orders of classes to be learned. 

\subsection{Wide application scenarios of the generative model}


\begin{figure}[!btp]
    \centering
    \includegraphics[height=0.37\linewidth]{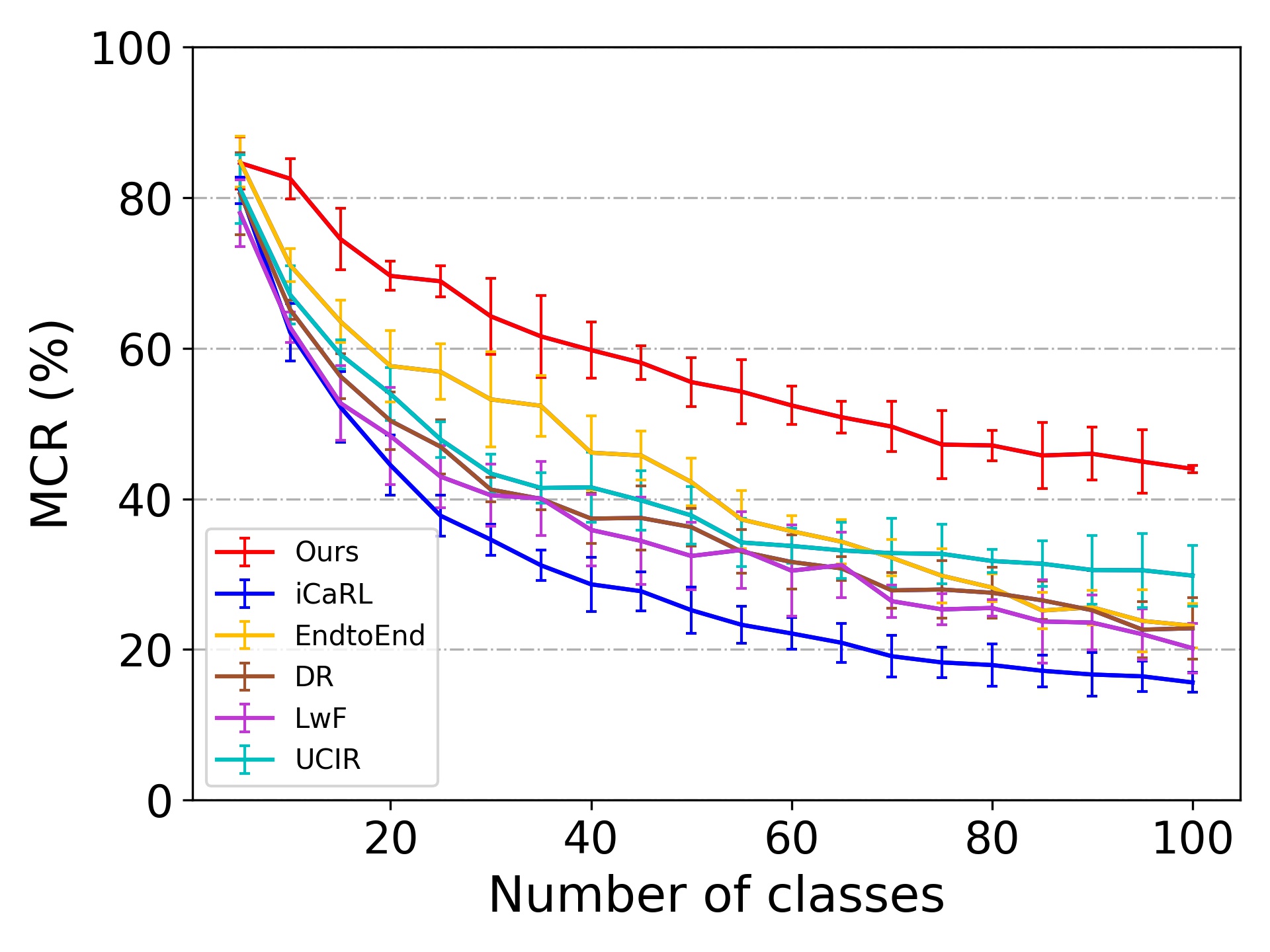}
    \includegraphics[height=0.37\linewidth]{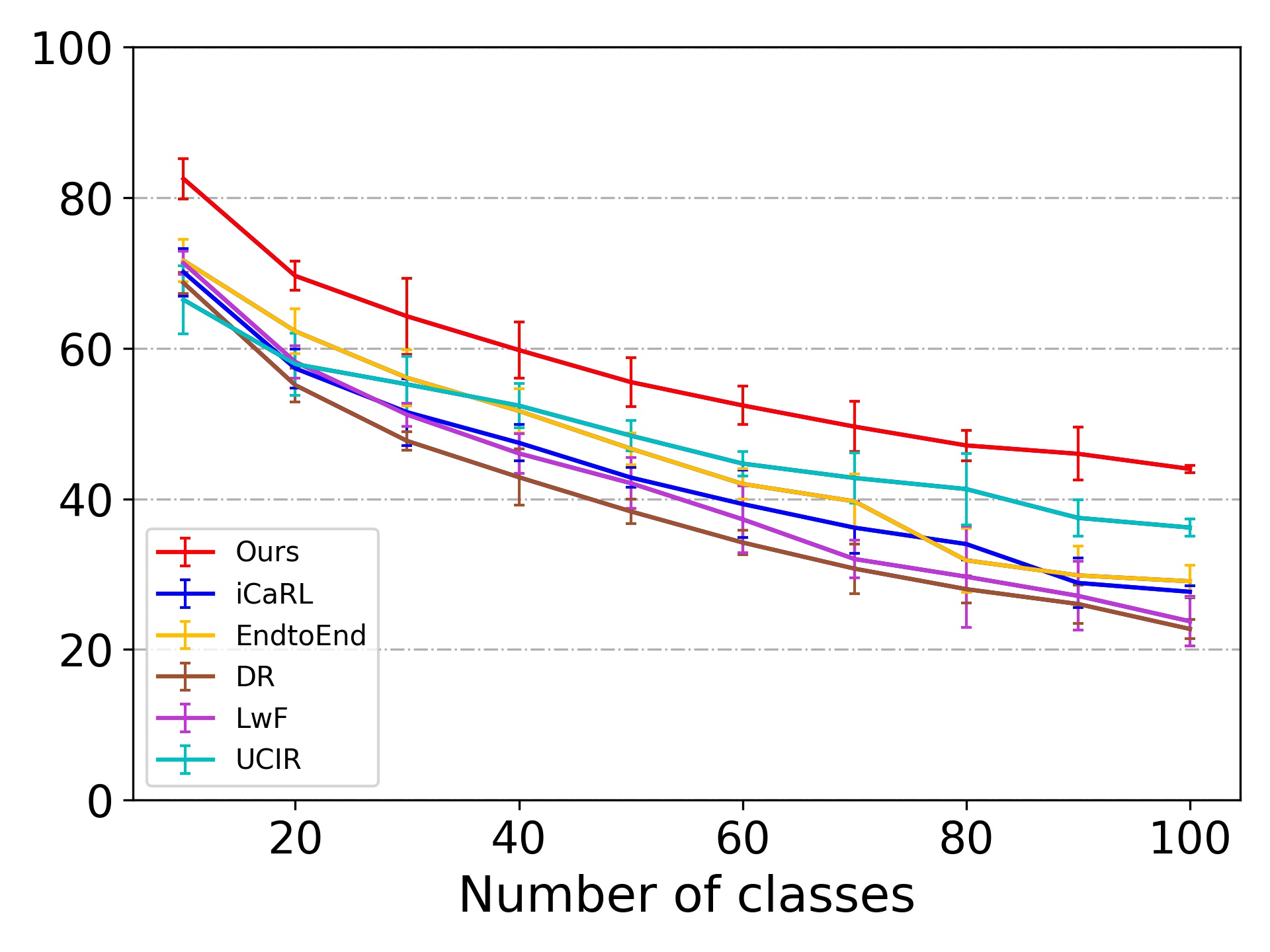}
    \includegraphics[height=0.37\linewidth]{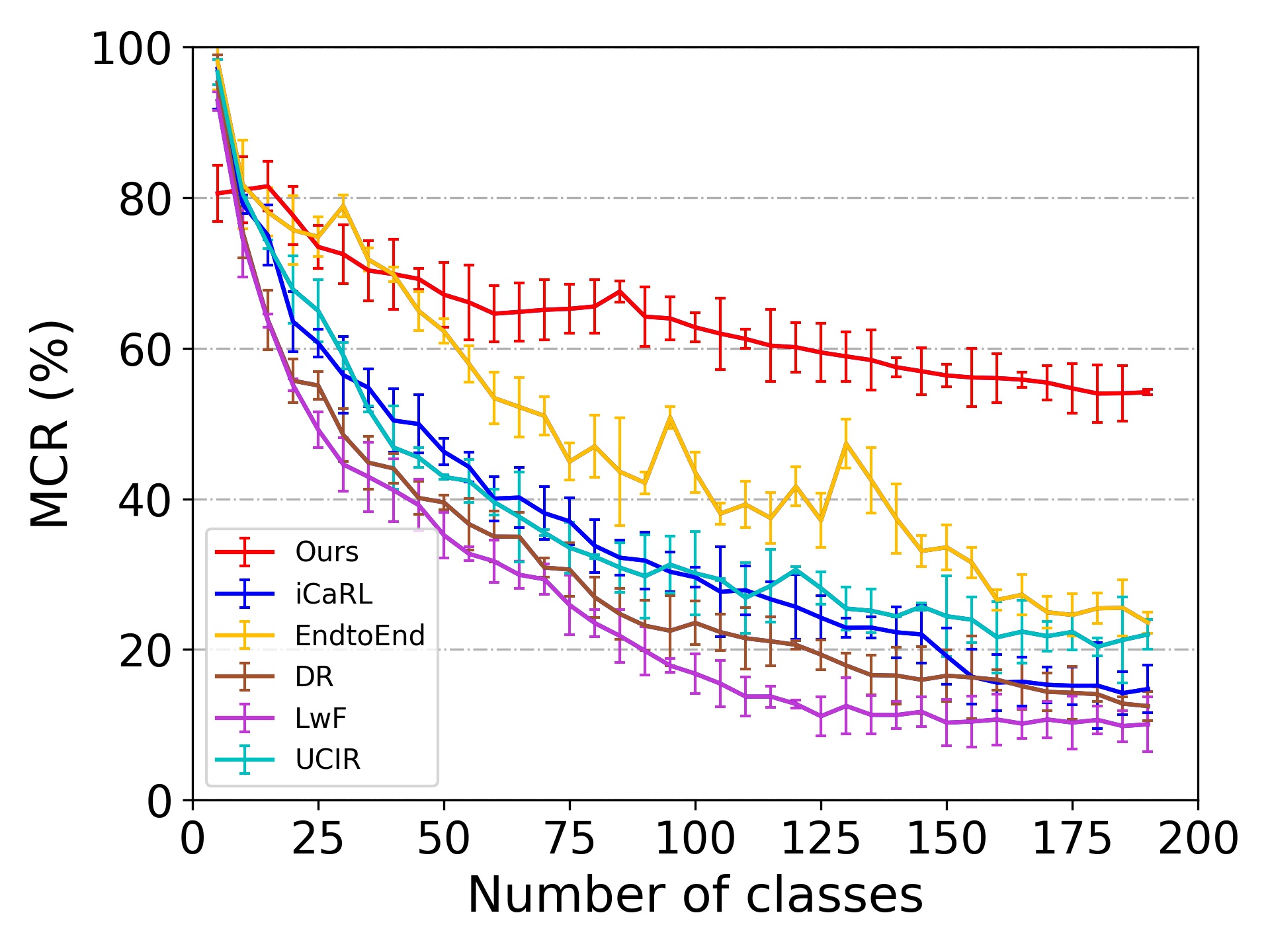}
    \includegraphics[height=0.37\linewidth]{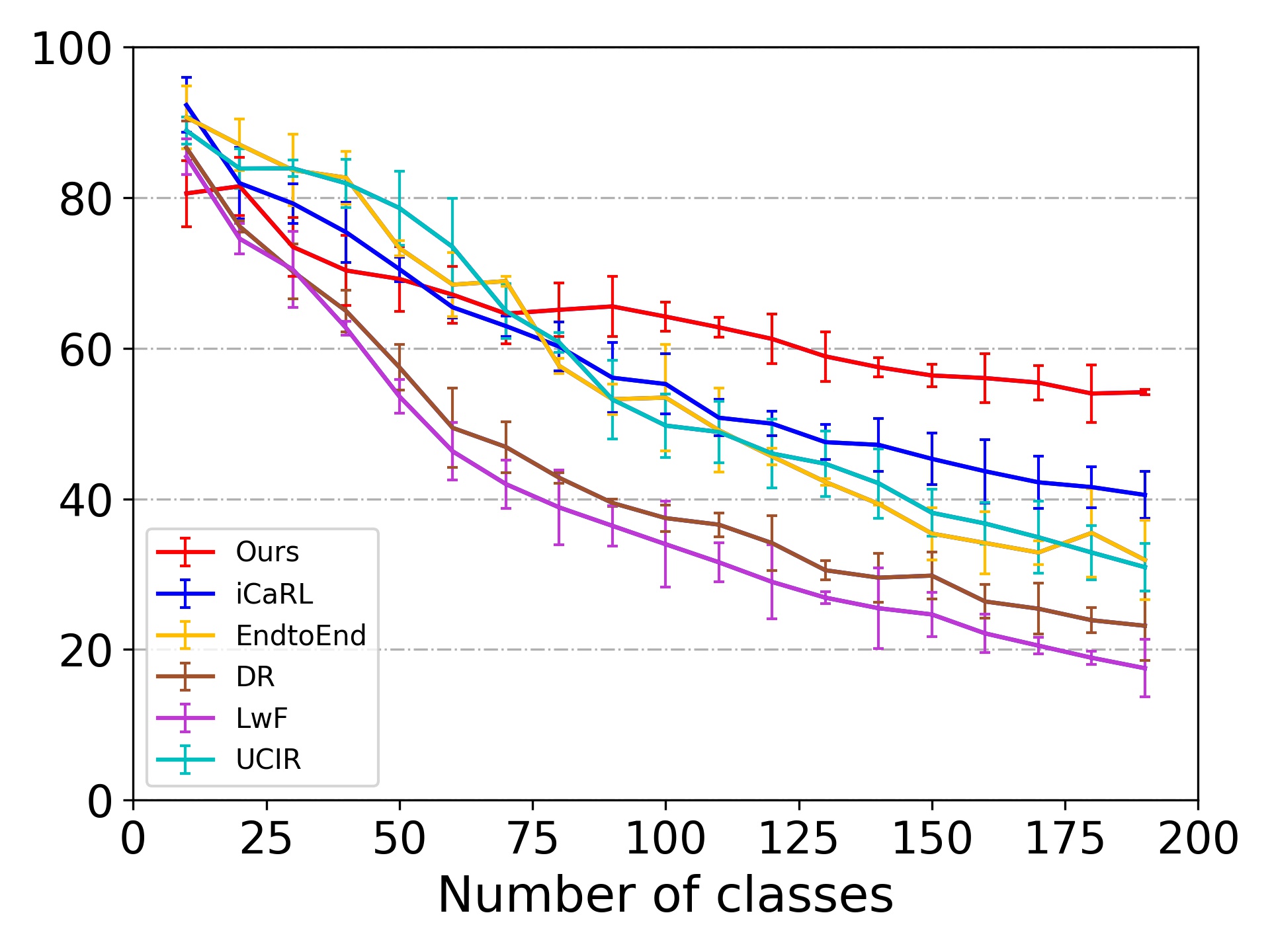}
    \caption{Few-shot continual learning performance on CIFAR100 and CUB200. Only 10 training images are available for each new class during continual learning on CIFAR100 (first row) and CUB200 (second row), with 5 (first column) or 10 (second column) new classes learned each time. }
    \label{fig:fewshot}
\end{figure}

One reason to explore continual learning techniques is due to the difficulty in collecting data from all classes. Such difficulty may cause another two challenging problems, few-shot continual learning where only a few number of training images are available for each new class at each round of continual learning, and data-incremental continual learning where the classifier would be updated continuously with new data of existing classes (rather than with data of new classes). To check whether the proposed approach works effectively in the scenario of few-shot continual learning, in the experiment, only 10 training images for each new class were provided to update the model for each method. The memory size was set 200 for all the baseline methods which need to store a small set of original images. Figure~\ref{fig:fewshot} shows that, while the strongest baseline method changes with varying dataset and number of learned new classes per round, the best performance is always from the proposed approach. This clearly supports that the proposed approach still works effectively even if only limited number of data is available during continual learning.   

\begin{figure}[!btp]
    \centering
    \includegraphics[height=0.37\linewidth]{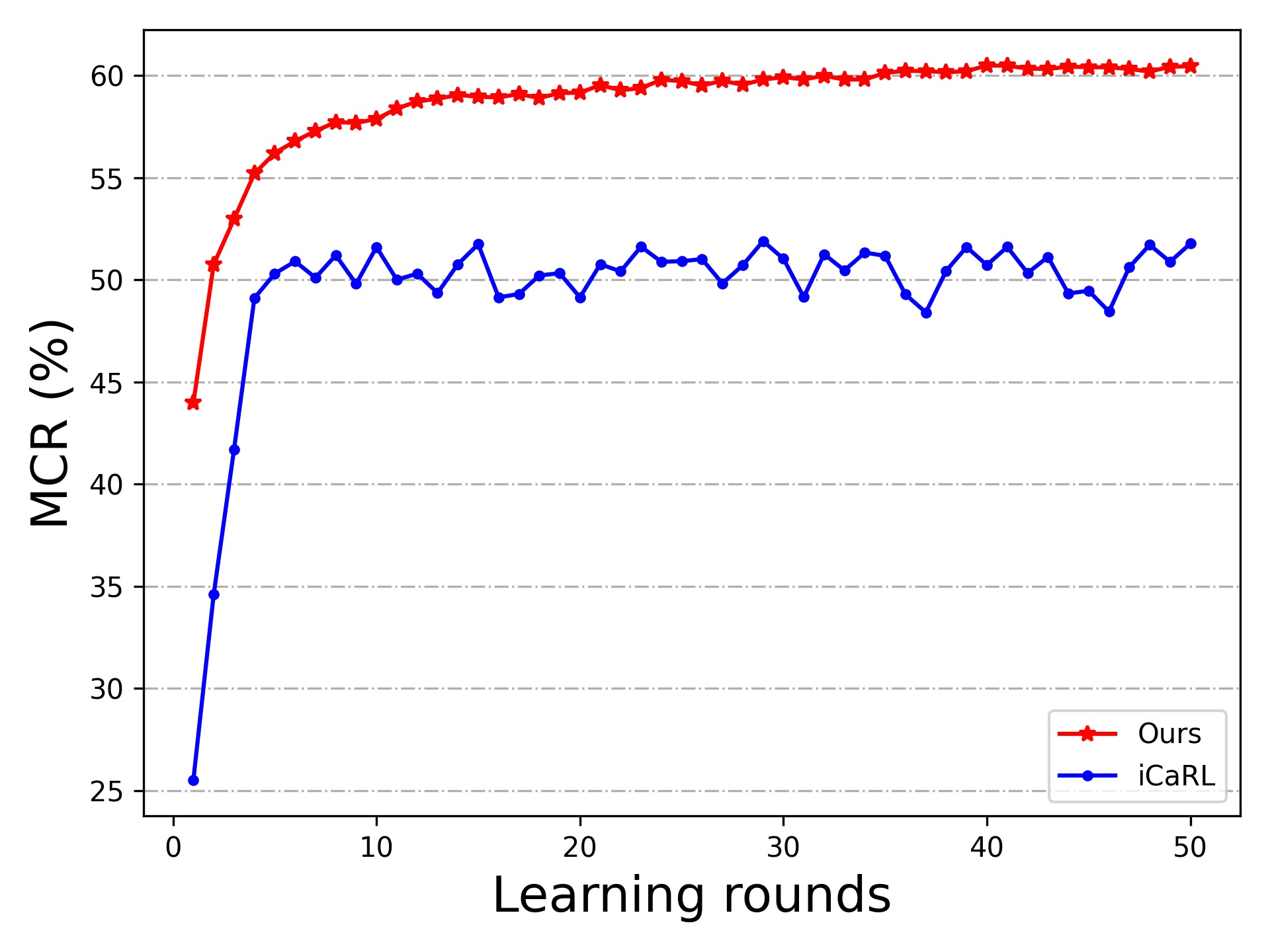}
    \includegraphics[height=0.37\linewidth]{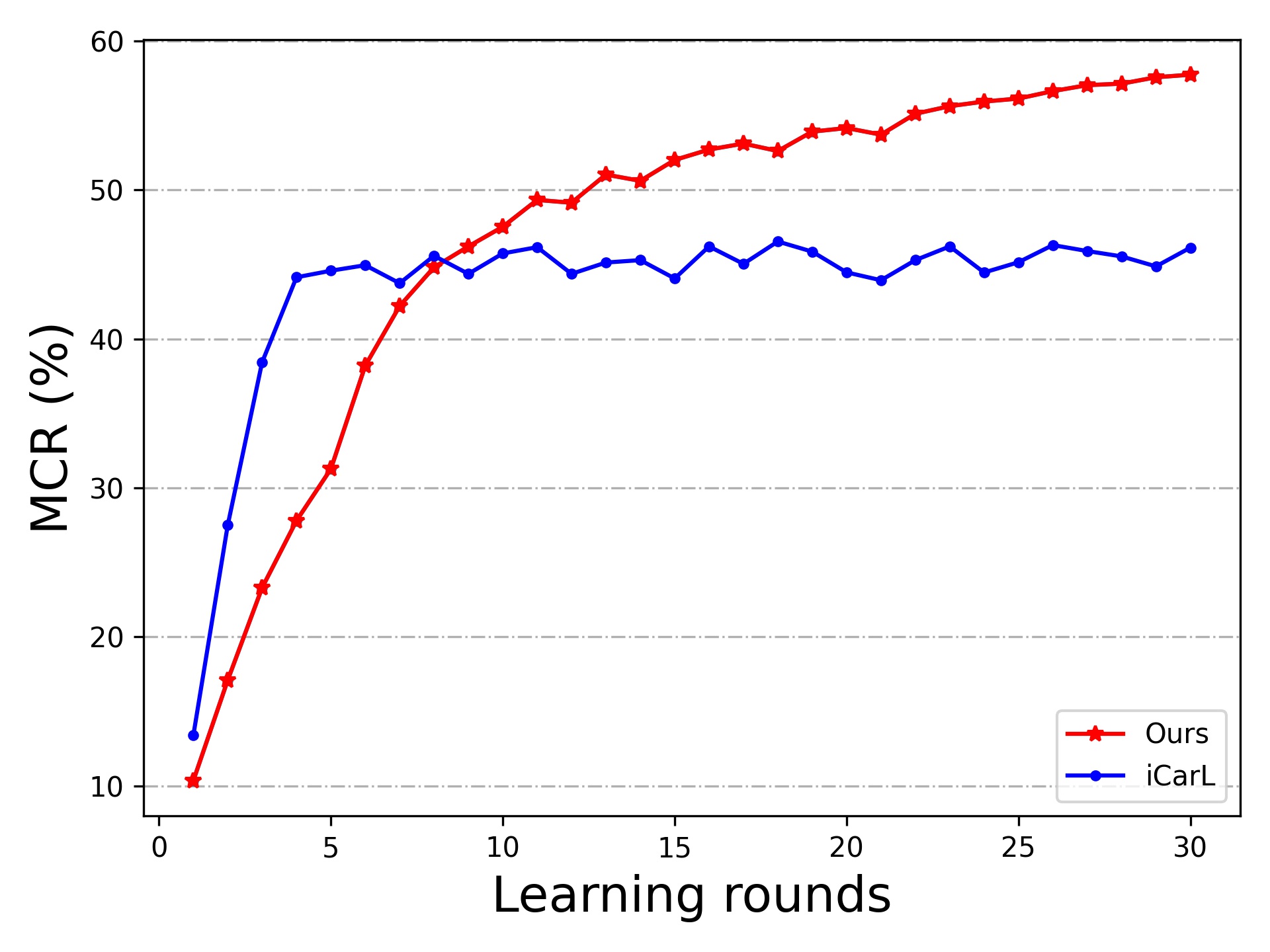}
    \caption{Data-incremental continual learning performance on CIFAR100 (Left) and CUB200 (Right). All classes are available from the beginning, but only 10 (CIFAR100) or 1 (CUB200) new images are available for each class at each round of continual learning. Memory size is respectively set 2000 (Left) and 400 (Right) for the representative method iCaRL. }
    \label{fig:data_incremental}
\end{figure}

For the data-incremental continual learning, with a few number of new images provided for each class at each round, it can be observed that the performance of the proposed approach increases over rounds of continual learning, while the  representative iCaRL method cannot effectively improve its learning performance shortly after the memory used in iCaRL becomes full (Figure~\ref{fig:data_incremental}). This is probably because the proposed approach can naturally update the representation of each class with more data, without discarding the information of previously appeared data. In comparison, the performance of the updated classifier by existing methods often depends on the limited original data stored in memory and the more recently appeared data. Note that the existing methods would perform even worse without  storing old data by memory.
This experiment demonstrates that the proposed approach can be used to handle two types (i.e., class-incremental and data-incremental) of continual learning, while existing methods can only handle the class-incremental learning task.

\subsection{Effect of feature extractor}
\begin{figure}[!btp]
    \centering
    \includegraphics[height=0.3\linewidth]{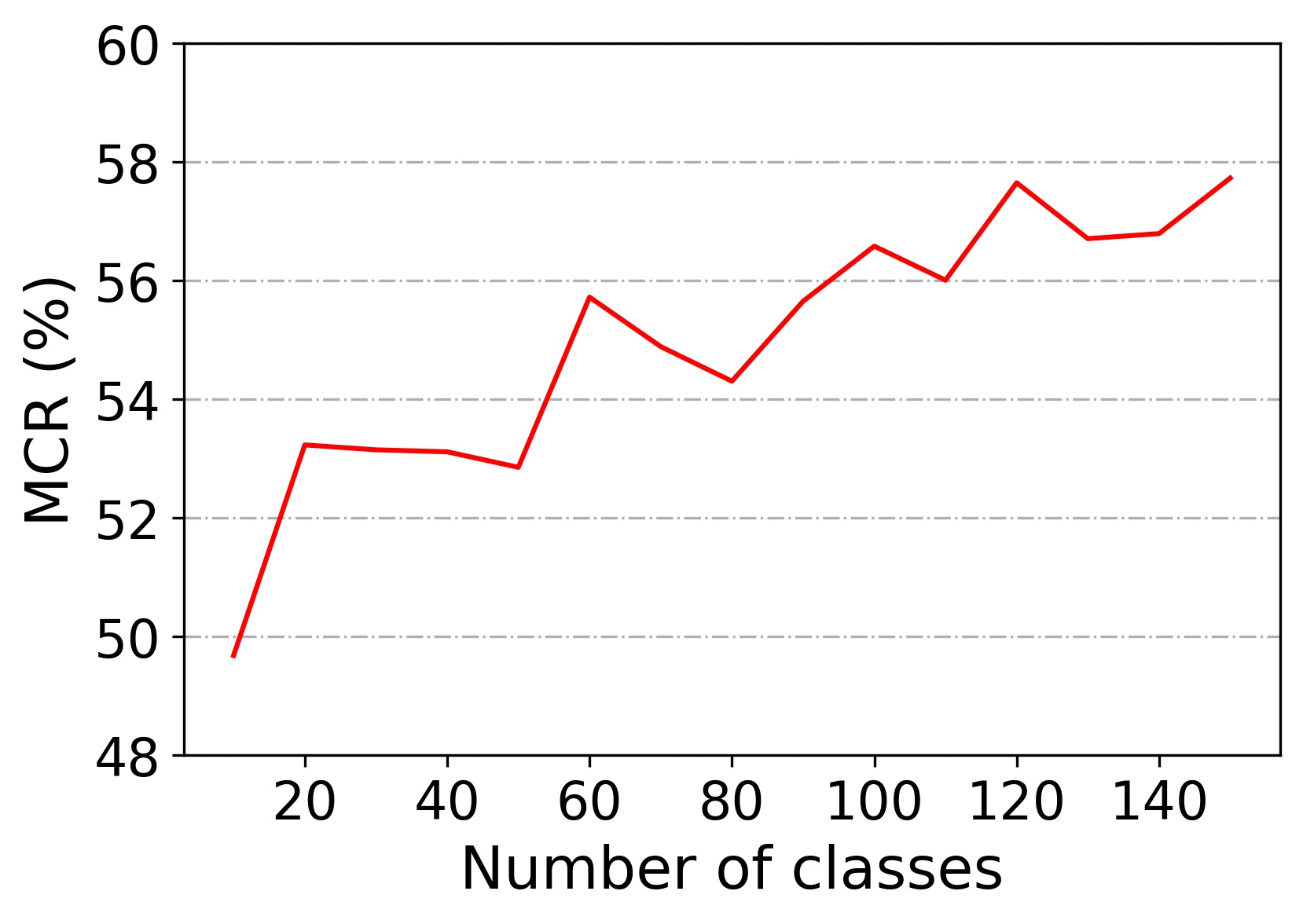} 
    \includegraphics[height=0.3\linewidth]{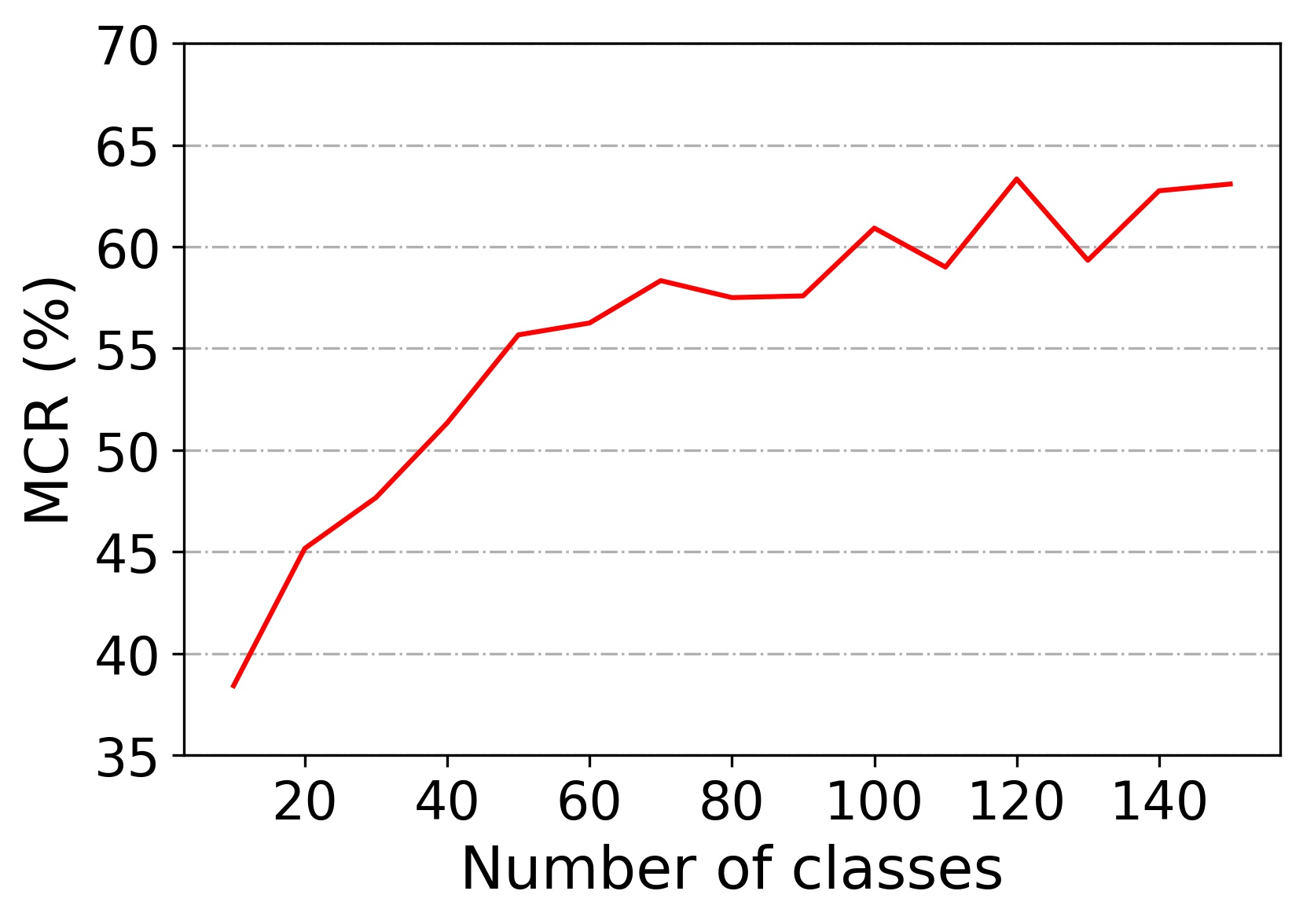}
    \caption{Effect of feature extractor on continual learning. More classes (x-axis) used to train feature extractor result in better performance on Skin7 (Left) and Skin40 (Right). Note that the classes used to train the feature extractor are not overlapped with the set of classes to be learned during continual learning.}
    \label{fig:featureextractor}
\end{figure}

The proposed approach is based on a fixed pre-trained feature extractor. To confirm that better feature extractors would help the generative model perform better in continual learning, the original 153 classes of skin image data used for training the feature extractor (before starting to continually learn new skin disease classes) were reduced gradually to only 10 classes, each time using such reduced number of classes to train the feature extractor and then the performance of the proposed approach at last round of continual learning on both the Skin7 and Skin40 datasets was calculated. Figure~\ref{fig:featureextractor} does show that more classes used for training the feature extractor would generally result in better performance of the proposed approach. The feature extractor trained by more classes of data would probably have learned to extract more types of features and therefore could be more generalizable to a new but relevant domain. Consistent with the observation and explanation, when the feature extractor is fixed by random parameter weights (i.e., without any training), the classifier in continual learning showed the worst performance (MCR is $21\%$ on Skin7, $6\%$ on Skin40; not shown in Figure~\ref{fig:featureextractor}). These results strongly suggest that exploring better ways to obtain a better feature extractor would further improve performance of the generative model in continual learning.

\begin{figure}[!btp]
    \centering
    \includegraphics[height=0.3\linewidth]{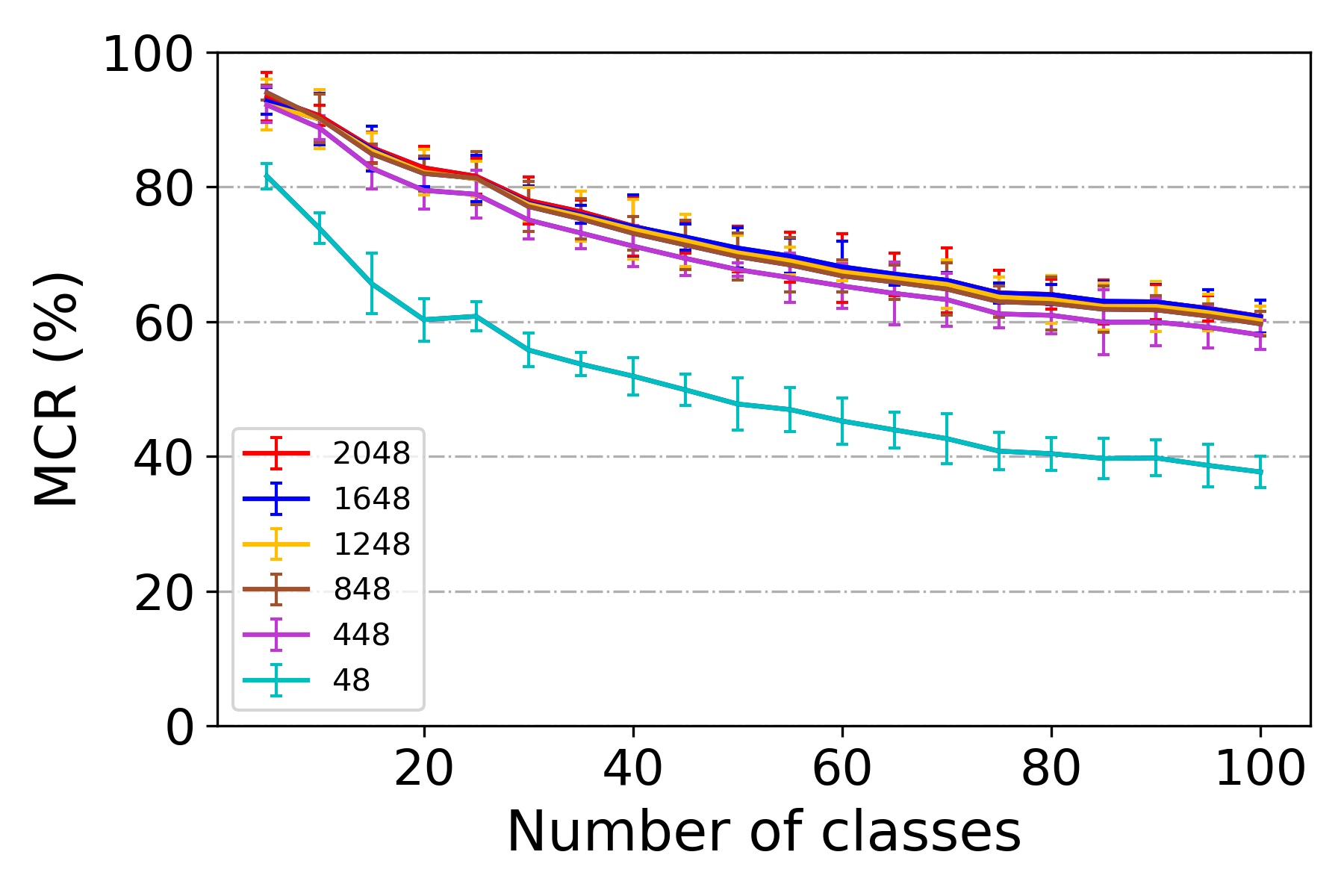}
    \includegraphics[height=0.3\linewidth]{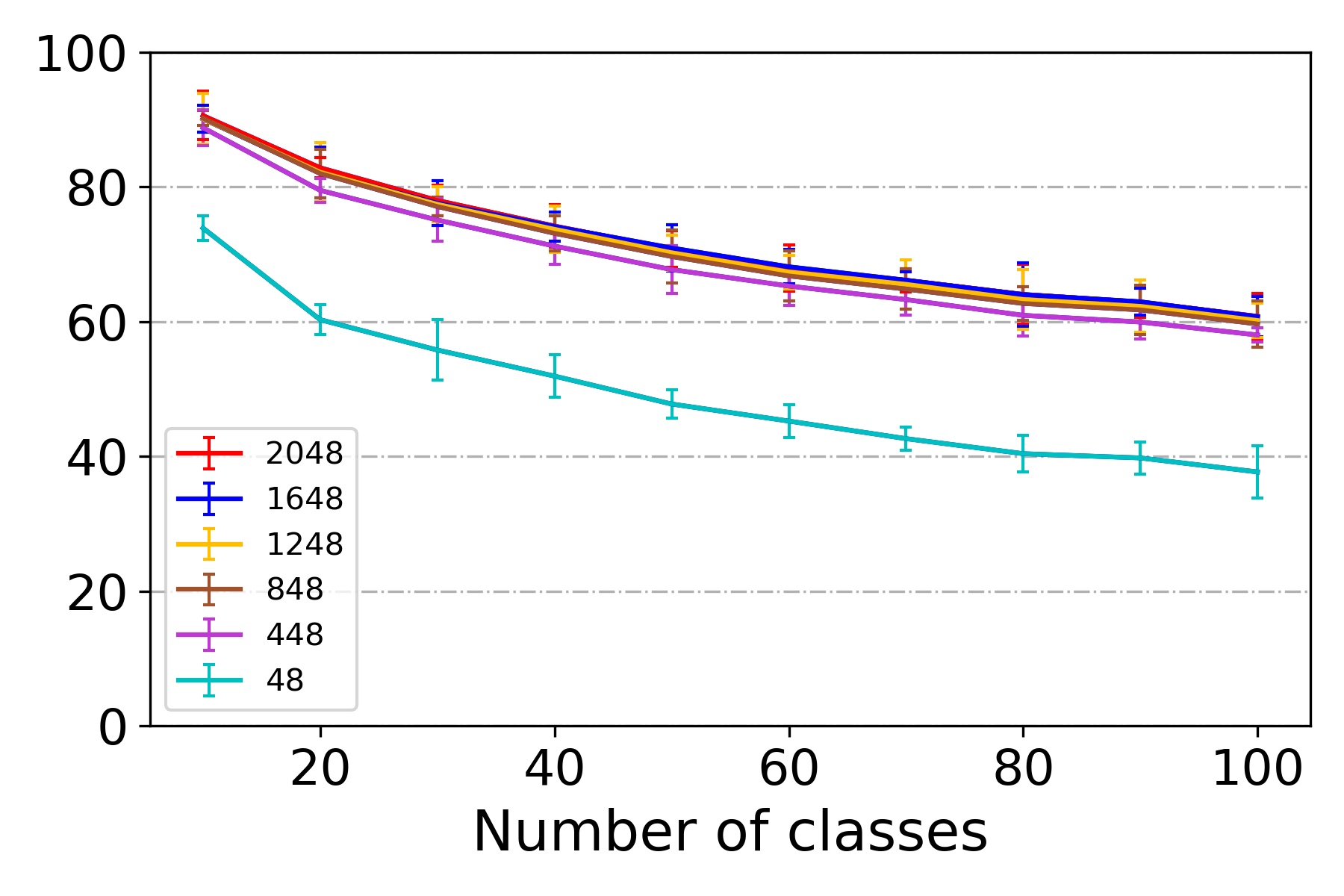}
    \includegraphics[height=0.3\linewidth]{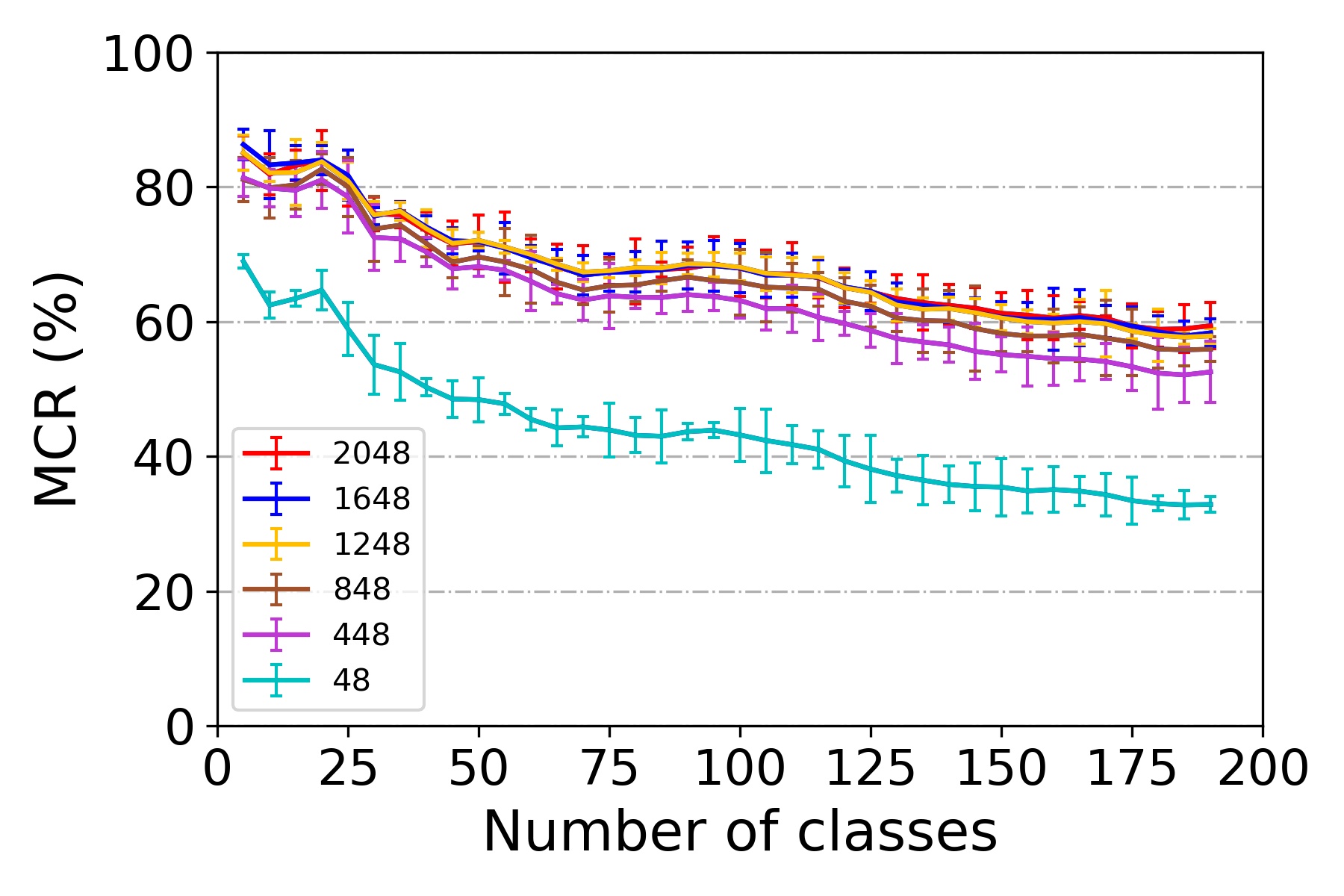}
    \includegraphics[height=0.3\linewidth]{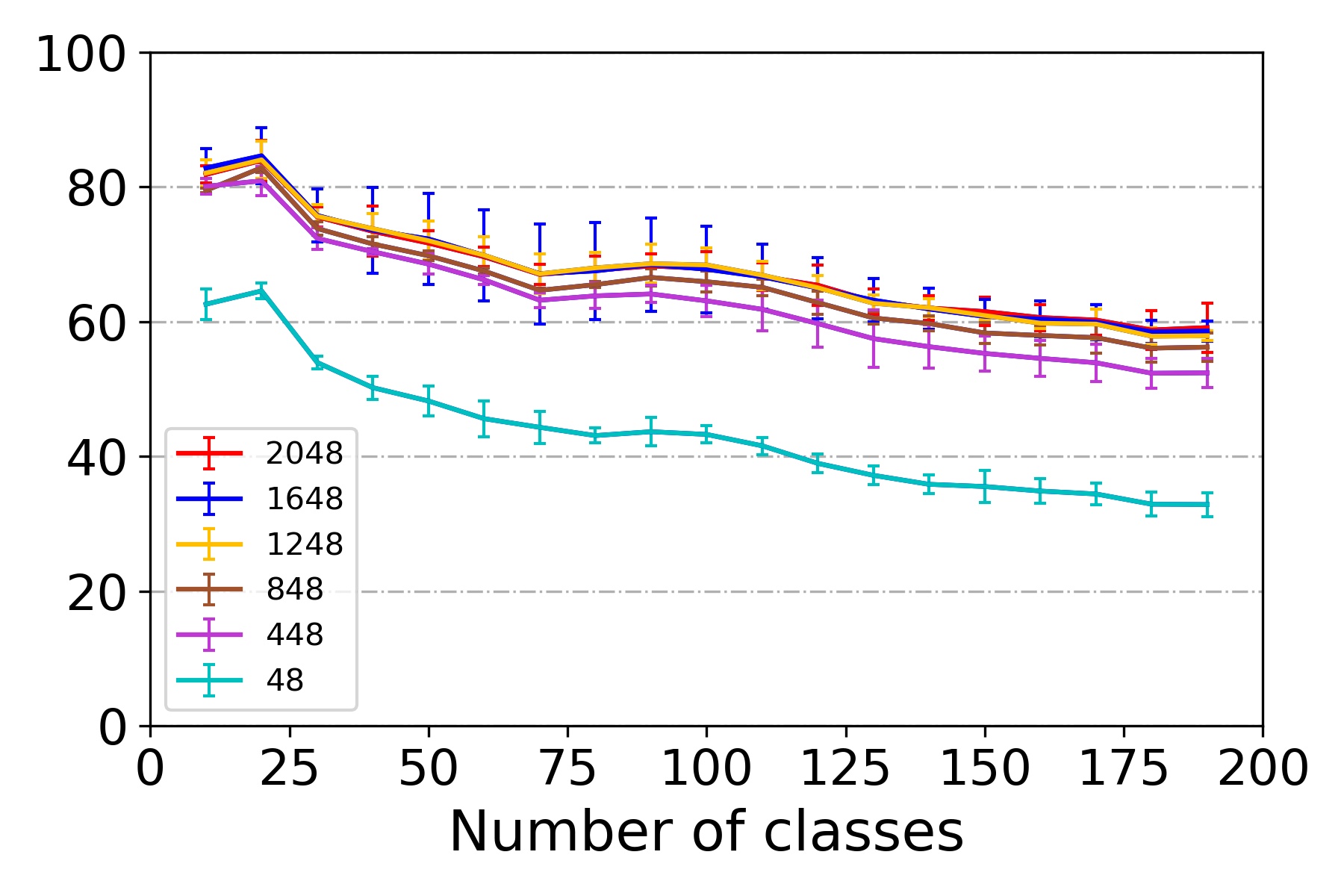}
    \caption{Continual learning performance with different numbers of feature outputs. First row: learning 5 (Left) and 10 (Right) new classes at each round on CIFAR100. Second row: learning 5 (Left) and 10 (Right) classes at each round on CUB200. Curves in different colors correspond to different numbers of feature outputs during continual learning.}
    \label{fig:featuredim}
\end{figure}

Another factor in the feature extractor to potentially affect continual learning performance is the size of feature extractor outputs. In general, more outputs could represent more types of feature information and therefore help the proposed approach represent richer information in each class. To confirm this hypothesis, varying number of feature outputs were randomly sampled from the original 2048 outputs, and then continual learning was performed based on the randomly sampled feature outputs. Figure~\ref{fig:featuredim} shows that the performance clearly downgrades when the used  feature extractor outputs are fewer than 1000, with more drops in performance corresponding to fewer outputs. Interestingly, the performance changes little when the number of feature extractor outputs decreases from the original 2048 to 1248. This may be because some outputs are highly correlated, such that removal of some of the outputs would not affect the representation power by the remaining outputs. This provides an opportunity to use relatively smaller number of outputs for knowledge representation when memory resource is very limited.

\section{Conclusion}
In this study, we propose a Bayesian generative model for continual learning of new classes. The model does not update the feature extractor but generates statistical information to represent knowledge of each class. Without storing any original data, the generative model can keep knowledge of each old class from forgetting and outperforms existing state-of-the-art approaches which often store small number of old data. The model is not limited to any specific feature extractor backbone and the ways to represent statistical information, and the final-round performance is not affected by the process of continual learning such as the number of new classes to be learned each time or the number of rounds of continual learning. Besides continually learning new classes, the model can also consistently improve the classification performance by continuously learning from new data of existing classes.
This study suggests a new direction to solve the catastrophic forgetting issue in continual learning, i.e., exploring effective ways to represent knowledge based on certain fixed but powerful pre-trained feature extractor. Better pre-trained feature extractor could also be explored to further improve the performance of the generative approach.

\section*{Acknowledgment}
This work is supported in part by the National Natural Science Foundation of China (grant No. 62071502, U1811461), and the Guangdong Key Research and Development Program (grant No. 2020B1111190001).

\bibliographystyle{IEEEbib}
\bibliography{ref}

\end{document}